%% file: arxiv_paper_v1.tex
\documentclass{article}
\usepackage{nips07submit_e,times}

\usepackage[dvipdfm]{graphicx}

\usepackage{amsmath}
\usepackage{amsfonts}
\usepackage{amssymb}
\usepackage{amsbsy} 
\usepackage{mathrsfs} 

\usepackage{amsthm}
\newtheorem{thm}{Theorem}
\newtheorem{prop}{Proposition}
\newtheorem{lem}{Lemma}
\newtheorem{cor}{Corollary}
\theoremstyle{definition} 
\theoremstyle{theorem} \newtheorem*{claim}{Claim}

\theoremstyle{definition} \newtheorem{example}{Example}

\newcommand{\ed}[2]{#1 \rightarrow #2}
\newcommand{\pd}[2]{\frac{\partial #1}{\partial #2}}
\newcommand{\pds}[3]{\frac{\partial^2 #1}{\partial #2 \partial #3}}
\newcommand{\spec}[1]{{\rm Spec}(#1)}

\newcommand{\covariance}[3]{{\rm Cov}_{#1}[#2,#3]}
\newcommand{\variance}[2]{{\rm Var}_{#1}[#2]}
\newcommand{\ev}[1]{\hat{#1}}
\newcommand{\nar}[1]{\hspace{-0.4mm} #1 \hspace{-0.4mm}}
\newcommand{\cg}[1]{\hat{#1}}

\DeclareMathOperator*{\argmin}{argmin}
\DeclareMathOperator*{\sgn}{sgn}

\newcommand{\es}{\hspace{-1.6mm}}  

\title{Graph Zeta Function in the Bethe Free Energy and Loopy Belief Propagation}

\author{
Yusuke Watanabe \\
The Institute of Statistical Mathematics\\
10-3 Midori-cho, Tachikawa\\
Tokyo 190-8562, Japan\\
\texttt{watay@ism.ac.jp} 
\And
Kenji Fukumizu \\
The Institute of Statistical Mathematics\\
10-3 Midori-cho, Tachikawa\\
Tokyo 190-8562, Japan\\
\texttt{fukumizu@ism.ac.jp} \\
}

\begin{document}

\maketitle

\begin{abstract}
We propose a new approach to the analysis of
Loopy Belief Propagation (LBP) by
establishing a formula that connects the Hessian of the Bethe free energy
with the edge zeta function.
The formula has a number of theoretical implications on LBP.  
It is applied to give a sufficient condition that the Hessian of
the Bethe free energy is positive definite,
which shows non-convexity for graphs with multiple cycles.
The formula clarifies the relation between the local stability of a fixed point of 
LBP and local minima of the Bethe free energy.
We also propose a new approach to the uniqueness of LBP fixed point,
and show various conditions of uniqueness.
\end{abstract}

\section{Introduction}

Pearl's belief propagation \cite{Pearl} provides an efficient method for
exact computation in the inference with probabilistic models
associated to trees. As an extension to general graphs allowing
cycles, Loopy Belief Propagation (LBP) algorithm
\cite{MWJempiricalstudy} has been proposed, showing successful
performance in various problems such as computer vision and error
correcting codes.

One of the interesting theoretical aspects of LBP is its connection
with the Bethe free energy \cite{YFWGBP}. It is known, for example,
the fixed points of LBP correspond to the stationary points of the
Bethe free energy. Nonetheless, many of the properties of LBP such
as exactness, convergence and stability are still unclear, and
further theoretical understanding is needed.

This paper theoretically analyzes LBP by establishing a formula
asserting that the determinant of the Hessian of the Bethe free
energy equals the reciprocal of the edge zeta function up to a
positive factor. This formula derives a variety of results
on the properties of LBP such as stability and uniqueness, since the
zeta function has a direct link with the dynamics of LBP as we show.

The first application of the formula is the condition for the
positive definiteness of the Hessian of the Bethe free energy. 
The Bethe free energy is not necessarily convex, which causes
unfavorable behaviors of LBP such as oscillation and multiple fixed
points. Thus, clarifying the region where the Hessian is positive
definite is an importance problem.  
Unlike the previous approaches which consider the global structure of the
Bethe free energy such as \cite{W1loop,Huniquness}, 
we focus the local structure.  Namely, 
we provide a simple sufficient
condition that determines the positive definite region: if all the
correlation coefficients of the pseudomarginals are smaller than a 
value given by a characteristic of the graph, the Hessian is
positive definite. Additionally, we show that the Hessian always has
a negative eigenvalue around the boundary of the domain if the graph
has at least two cycles.

Second, we clarify a relation between the local stability of a LBP
fixed point and the local structure of the Bethe free energy. 
Such a relation is not necessarily obvious, since LBP is
not the gradient descent of the Bethe free energy.  
In this line of studies, Heskes \cite{Hstable} shows that a locally stable
fixed point of LBP is a local minimum of the Bethe free energy.  It
is thus interesting to ask which local minima of the Bethe free
energy are stable or unstable fixed points of LBP.   We answer this
question by elucidating the conditions of the local stability of LBP
and the positive definiteness of the Bethe free energy in terms of
the eigenvalues of a matrix, which appears in the graph zeta
function.
%

Finally, we discuss the uniqueness of LBP fixed point by developing
a differential topological result on the Bethe free energy. The result
shows that the determinant of the Hessian at the fixed points, which
appears in the formula of zeta function, must satisfy a strong
constraint. As a consequence, in addition to the known result on the
one-cycle case, we show that the LBP fixed point is unique for any
unattractive connected graph with two cycles
without restricting the strength of interactions.

\section{Loopy belief propagation algorithm and the Bethe free energy}
\label{sectionLBPandBethe}
Throughout this paper,
$G=(V,E)$ is a connected undirected graph with $V$
the vertices and $E$ the undirected edges.
The cardinality of $V$ and $E$ are denoted by $N$ and $M$
respectively.

In this article we focus on binary variables, {\it i.e.}, 
$x_i \in \{\pm 1\}$. 
Suppose that the probability distribution
over the set of variables
$\boldsymbol{x}=(x_i)_{i \in V}$
is given by the following factorization form
with respect to $G$:
\vspace{-1mm}
\begin{equation}
p(\boldsymbol{x}) =
\frac{1}{Z}
\prod_{ij \in E}\psi_{ij}(x_i,x_j)
\prod_{i \in V}\psi_{i}(x_i), \label{exactprobability}
\end{equation}
where $Z$ is a normalization constant and
$\psi_{ij} \text{ and } \psi_{i}$ are positive functions given by $\psi_{ij}(x_i,x_j)=\exp (J_{ij} x_i x_j)$
and $\psi_{i}(x_i)=\exp (h_i x_i)$
without loss of generality.

In various applications, the computation of marginal distributions
$p_{i}(x_i):=\sum_{\boldsymbol{x} \setminus \{x_i\}
}p(\boldsymbol{x})$ and $p_{ij}(x_i,x_j):=\sum_{\boldsymbol{x}
\setminus \{x_i x_j\} } p(\boldsymbol{x})$ is required though 
the exact computation is intractable for large graphs.
If the graph is a
tree, they are efficiently computed by Pearl's belief propagation
algorithm \cite{Pearl}. Even if the graph has cycles, it is
empirically known that the direct application of this algorithm, called
Loopy Belief Propagation (LBP), often gives good approximation.

LBP is a message passing algorithm.
For each directed edge, a message vector $\mu_{\ed{i}{j}}(x_j)$
is assigned and initialized arbitrarily.
The update rule of messages is given by
\begin{equation}
\mu^{{\rm new}}_{\ed{i}{j}}(x_j)
\propto
\sum_{x_i}
\psi_{ji}(x_j,x_i) \psi_{i}(x_i)
\prod_{k \in N_i \setminus j}
\mu_{\ed{k}{i}}(x_i),           \label{BPupdate}
\end{equation}
where $N_i$ is
the neighborhood of $i \in V$. 
The order of edges in
the update is arbitrary. In this paper we consider {\it parallel
update}, that is, all edges are updated simultaneously. If the
messages converge to a fixed point
$\{\mu^{\infty}_{\ed{i}{j}} \}$, 
the approximations of
$p_{i}(x_i)$ and $p_{ij}(x_i,x_j)$ are calculated by the beliefs,
\begin{small}
\begin{equation}
b_i(x_i) \propto  \psi_{i}(x_i)
\hspace{-1mm}
\prod_{k \in N_i} \mu^{\infty}_{\ed{k}{i}}(x_i),
\quad
b_{ij}(x_i,x_j) \propto \psi_{ij}(x_i,x_j) \psi_{i}(x_i) \psi_{j}(x_j)
\hspace{-2mm}
\prod_{k \in N_i \setminus j}\hspace{-1.5mm} \mu^{\infty}_{\ed{k}{i}}(x_i)
\hspace{-1mm}
\prod_{k \in N_j \setminus i}\hspace{-1.5mm}
\mu^{\infty}_{\ed{k}{j}}(x_j), \label{DefBelief} 
\end{equation}
\end{small}
\es
with normalization $\sum_{x_i}b_{i}(x_i)=1$
and $\sum_{x_i,x_j}b_{ij}(x_i,x_j)=1$.
From (\ref{BPupdate}) and (\ref{DefBelief}), the constraints
$b_{ij}(x_i,x_j) > 0$ and $\sum_{x_j}b_{ij}(x_i,x_j)=b_i(x_i)$ are automatically satisfied.
%

We introduce the Bethe free energy
as a tractable approximation of the Gibbs free energy.
The exact distribution (\ref{exactprobability})
is characterized by a variational problem
$p(\boldsymbol{x})= \argmin_{\hat{p}} F_{Gibbs}( \hat{p} )$,
where the minimum is taken over all probability distributions on
$(x_i)_{i \in V}$ and $F_{Gibbs}( \hat{p})$ is the 
{\it Gibbs free energy} defined by 
$F_{Gibbs}( \hat{p})=KL(\hat{p}||p)-\log Z$.  
Here $KL(\hat{p}||p)=\int\hat{p}\log(\hat{p}/p)$ is the Kullback-Leibler
divergence from $\hat{p}$ to $p$. 
Note that $F_{Gibbs}( \hat{p})$ is a convex function of $\hat{p}$.

In the Bethe approximation,
we confine the above minimization
to the distribution of the form $b( \boldsymbol{x} ) \propto
\prod_{ij \in E} b_{ij}(x_i,x_j) \prod_{i \in V} b_{i}(x_i)^{1-d_i}$
, where $d_i:=|N_i|$ is the degree and the constraints
$b_{ij}(x_i,x_j) > 0$,
$\sum_{x_i,x_j}b_{ij}(x_i,x_j)=1$ and
$\sum_{x_j}b_{ij}(x_i,x_j)=b_{i}(x_i)$
are satisfied.  A set $\{b_{i}(x_i),b_{ij}(x_i,x_j)\}$ satisfying
these constraints is called {\it pseudomarginals}. For
computational tractability, we modify the Gibbs free energy to the
objective function called {\it Bethe free energy}:
\begin{align}
F( b ):=
&-
\sum_{ij \in E}\sum_{x_i x_j}b_{ij}(x_i,x_j)\log{ \psi_{ij}(x_i,x_j) }
-
\sum_{i \in V}\sum_{x_i}b_{i}(x_i)\log{ \psi_{i}(x_i) } \nonumber \\
& \qquad +
\sum_{ij \in E}\sum_{x_i x_j}b_{ij}(x_i,x_j)\log{ b_{ij}(x_i,x_j) }
+
\sum_{i \in V} (1-d_i) \sum_{x_i}b_i(x_i)\log{b_i(x_i)}. \label{BetheFreeEnergy}
\end{align}
%
The domain of the objective function $F$ is the set of
pseudomarginals. The function $F$ does not necessarily have 
a unique minimum.
The outcome of this modified variational problem is the same as
that of LBP \cite{YFWGBP}. To put it more precisely,
There is a one-to-one correspondence between the set of stationary
points of the Bethe free energy
and the set of fixed points of LBP.

It is more convenient if we work with minimal parameters, mean $m_i=
\text{E}_{b_i}[x_i]$ and correlation $\chi_{ij}=
\text{E}_{b_{ij}}[x_i x_j]$. Then we have an effective
parametrization of pseudomarginals:
\vspace{-1.0mm}
\begin{equation}
b_{ij}(x_i,x_j)=\frac{1}{4}(1+m_i x_i + m_j x_j + \chi_{ij}x_i x_j ),
\quad \quad
b_{i}(x_i)=\frac{1}{2}(1+m_i). \label{defbbymchi}
\end{equation}
The Bethe free energy (\ref{BetheFreeEnergy}) is rewritten as
\begin{align}
F&(\{m_i,\chi_{ij}\})=
-
\sum_{ij \in E}J_{ij}\chi_{ij}
-
\sum_{i \in V}h_i m_i \nonumber \\
&+
\sum_{ij \in E}\sum_{x_i x_j}
\eta \Big(\frac{\nar{1}+\nar{m_i x_i} +\nar{ m_j x_j }+
 \chi_{ij} x_i x_j}{4} \Big)
+
\sum_{i \in V} (1-d_i) \sum_{x_i}
 \eta \Big(\frac{1+\nar{m_i x_i}}{2} \Big),
 \label{BetheFreeEnergybinary}
\vspace{-1mm}
\end{align}
where $\eta (x):=x \log x$. The domain of $F$ is written as
\begin{equation}
L(G):=
\Big\{
\{m_i,\chi_{ij}\} \in \mathbb{R}^{N+M}|
1+m_i x_i + m_j x_j + \chi_{ij}x_i x_j  > 0
\text{   for all } ij \in E \text{ and } x_i,x_j= \pm 1
\Big\}.\nonumber
\end{equation}

The Hessian of $F$, which consists of the second derivatives with
respect to $\{m_i,\chi_{ij}\}$,
is a square matrix of size $N+M$ and
denoted by $\nabla^2 F$.
This is considered to be a matrix-valued function on $L(G)$.
Note that, from (\ref{BetheFreeEnergybinary}), $\nabla^2 F$ does not depend on
$J_{ij}$ and $h_i$.

\section{Zeta function and Hessian of Bethe free energy} \label{secmainformula}

\subsection{Zeta function and Ihara's formula}
For each undirected edge of $G$, we make a pair of oppositely
directed edges, which form a set of {\it directed edges} $\vec{E}$.
Thus $|\vec{E}|=2M$. For each directed edge $e \in \vec{E}$, $o(e)
\in V$ is the {\it origin} of $e$ and $t(e) \in V$ is the {\it
terminus} of $e$.   For $e \in \vec{E}$, the {\it inverse edge} is
denoted by $\bar{e}$, and the corresponding undirected edge by
$[e]=[\bar{e}] \in E$.

A {\it closed geodesic} in $G$ is a sequence $(e_1,\ldots,e_k)$ of
directed edges such that 
$t(e_i)=o(e_{i+1}) \text{ and } e_i \neq \bar{e}_{i+1}$
for $i \in \mathbb{Z}/k\mathbb{Z}$.
Two closed geodesics are said to be {\it equivalent} if
one is obtained by cyclic permutation of the other.
An equivalent class of closed geodesics is called a {\it prime cycle}
if it is not a repeated concatenation of a shorter closed geodesic.
Let $P$ be the set of prime cycles of $G$.
For given weights $\boldsymbol{u}=(u_e)_{e \in \vec{E}}$,
the {\it edge zeta function}
\cite{Hpadic,STzeta1} is defined by
\begin{equation}
\zeta_{G}(\boldsymbol{u}):=\prod_{\mathfrak{p} \in P }
(1-g(\mathfrak{p}))^{-1},
\quad
g(\mathfrak{p}):=u_{e_1} \cdots u_{e_k}
\quad
\text{ for }
\mathfrak{p}=(e_1,\ldots,e_k), \nonumber
\end{equation}
where $u_{e} \in \mathbb{C}$ is assumed to be sufficiently small for
convergence.
This is an analogue of the Riemann zeta function which is represented by the
product over all the prime numbers.

\begin{example} \label{example1}
If $G$ is a tree, which has no prime cycles,
$\zeta_{G}(\boldsymbol{u})^{}=1$. For 1-cycle graph $C_N$ of length
$N$, the prime cycles are $(e_1,e_2,\ldots,e_N)$ and
$(\bar{e}_N,\bar{e}_{N-1},\ldots,\bar{e}_1)$, and thus $\zeta_{C_N}(\boldsymbol{u})=
(1-\prod_{l=1}^{N}u_{e_l})^{-1}
(1-\prod_{l=1}^{N}u_{\bar{e}_l})^{-1}.$ Except for these two types
of graphs, the number of prime cycles is infinite.
\end{example}

It is known that the edge zeta function has the following simple
determinant formula, which gives analytical continuation to the
whole $\mathbb{C}^{2M}$. Let $C(\vec{E})$ be the set of functions on
the directed edges. We define a matrix on $C(\vec{E})$, which is
determined by the graph $G$, by
\begin{equation}
\mathcal{M}_{e,e'}:=
\begin{cases}
1  \qquad \text{if } e \neq \bar{e'} \text{ and } o(e)=t(e'), \\
0  \qquad \text{otherwise.}
\end{cases}  \label{DirectedEdgeMatrix}
\end{equation}

\begin{thm}
[\cite{STzeta1}, Theorem 3]
\begin{equation}
\zeta_{G}(\boldsymbol{u})=\det(I- \mathcal{U}\mathcal{M})^{-1},
\end{equation}
where $\mathcal{U}$ is a diagonal matrix defined by
$\mathcal{U}_{e,e'}:=u_e \delta_{e,e'}$.
\end{thm}

%
%

We need to show another determinant formula of the edge zeta function,
which is used in the proof of Theorem \ref{thmmain}.
We leave the proof of Theorem \ref{thmEdgeIharaFormula} to the
supplementary material.
%
\begin{thm}
[Multivariable version of Ihara's formula] \label{thmEdgeIharaFormula}
Let $C(V)$ be the set of functions on $V$.
We define two linear operators on $C(V)$ by
\begin{equation}
 (\mathcal{\ev{D}}f)(i):=
\Big(
\sum_{e \in \vec{E} \atop t(e)=i}\frac{u_e u_{\bar{e}}}{1- u_e
u_{\bar{e}}}
\Big)
f(i),
\quad
 (\mathcal{\ev{A}}f)(i):=
\sum_{e \in \vec{E} \atop t(e)=i}
\frac{u_e}{1-u_e u_{\bar{e}}} f(o(e)),
\quad
\text{ where } f \in C(V). \label{defofDprimeAprime}
\end{equation}
Then we have
\vspace{-0.5mm}
\begin{equation}
\Big(
\zeta_{G}(\boldsymbol{u})^{-1}
=
\Big)
\det(I- \mathcal{U}\mathcal{M})
=
\det(I+ \mathcal{\ev{D}}-\mathcal{\ev{A}})
\prod_{[e]\in E}(1-u_e u_{\bar{e}}).
\label{EdgeIharaFormula}
\end{equation}
\end{thm}
%
%
%
%
%
If we set $u_e=u$ for all $e \in \vec{E}$
, the edge zeta function is called
the {\it Ihara zeta function} \cite{Idiscrete}
and denoted by $\zeta_{G}(u)$.
In this single variable case,
Theorem \ref{thmEdgeIharaFormula} is
reduced to Ihara's formula \cite{Bass}:
\vspace{-0.5mm}
\begin{equation}
\zeta_{G}(u)^{-1}
=
\det(I- u \mathcal{M})
=(1-u^2)^{M}
\det(I+\frac{u^2}{1-u^2}\mathcal{D}-\frac{u}{1-u^2}\mathcal{A} ), \label{IharaFormula}
\end{equation}
where
$\mathcal{D}$ is the {\it degree matrix}
and
$\mathcal{A}$ is the {\it adjacency matrix}
defined by
\begin{equation}
(\mathcal{D}f)(i):=
d_i f(i),
\quad
(\mathcal{A}f)(i):=
\sum_{e \in \vec{E} ,  t(e)=i}
f(o(e)),
\quad
\text{    }
f \in C(V). \nonumber
\end{equation}
%
%
\subsection{Main formula} 
\begin{thm}[Main Formula]
 \label{thmmain}
The following equality holds at any point of $L(G)$:
\begin{equation}
\Big(
\zeta_{G}(\boldsymbol{u})^{-1}
\hspace{-1mm}
=
\hspace{-1mm}
\Big)
\det(I- \mathcal{U}\mathcal{M})
=
\det(\nabla^2 F)
\prod_{ij \in E}\prod_{x_i, x_j = \pm 1}
\hspace{-2.5mm}
b_{ij}(x_i,x_j)
\prod_{i \in V}\prod_{x_i = \pm 1}
\hspace{-1mm}
b_{i}(x_i)^{1-d_i}
\text{ }
2^{2N+4M},  \label{eqthmmain}
\end{equation}
where
$b_{ij}$ and $b_i$ are given by (\ref{defbbymchi}) and
\vspace*{-1mm}
\begin{equation}
u_{\ed{i}{j}}:= \frac{\chi_{ij}-m_i m_j}{1-m_j^2}. \label{defuij}
\end{equation}
\end{thm}

\vspace*{-1mm}
\begin{proof}
(The detail of the computation is given in the supplementary material.)
\\
From ($\ref{BetheFreeEnergybinary}$),
it is easy to see that
the (E,E)-block of the Hessian is a diagonal matrix given by
\begin{small}
\begin{equation}
\pds{F}{\chi_{ij}}{\chi_{kl}}=
\delta_{ij,kl}
\frac{1}{4}
\Big(
\frac{1}{\nar{1}+\nar{m_i}+\nar{m_j}+\nar{\chi_{ij}}}
+\frac{1}{\nar{1}-\nar{m_i}+\nar{m_j}-\nar{\chi_{ij}}}
+\frac{1}{\nar{1}+\nar{m_i}-\nar{m_j}-\nar{\chi_{ij}}}
+\frac{1}{\nar{1}-\nar{m_i}-\nar{m_j}+\nar{\chi_{ij}}}
\Big).   \nonumber
\end{equation}
\end{small}
Using this diagonal block, we erase (V,E)-block and (E,V)-block of
the Hessian. In other words, we choose a square matrix $X$ such that
$\det X =1$ and
\begin{equation}
X^T (\nabla^2 F) X
=
\begin{bmatrix}
\quad Y & 0 \\
\quad 0 &
\Big( \pds{F}{\chi_{ij}}{\chi_{kl}} \Big)
\end{bmatrix}.  \nonumber
\end{equation}
After the computation given in the supplementary material, we see that
\begin{equation}
(Y)_{i,j}
=
\begin{cases}
\frac{1}{1-m_i^2}
+\sum_{k \in N_i}\frac{(\chi_{ik}-m_i m_k)^2}
{(1-m_i^2)(1-m_i^2-m_k^2+2 m_i m_k \chi_{ik}-\chi_{ik}^2)}
&\text{ if } i=j,  \\
- \mathcal{A}_{i,j}
\frac{\chi_{ij}-m_i m_j}{1-m_i^2-m_j^2+2 m_i m_j \chi_{ij}-\chi_{ij}^2}
&\text{ otherwise. } \label{eqYformula}
\end{cases}
\end{equation}
From $u_{\ed{j}{i}}= \frac{\chi_{ij}-m_i m_j}{1-m_i^2}$,
it is easy to check that
$I_N + \mathcal{\ev{D}} - \mathcal{\ev{A}} = Y W$,
where $\mathcal{\ev{A}}$ and $\mathcal{\ev{D}}$ is defined in
(\ref{defofDprimeAprime}) and
$W$ is a diagonal matrix defined by $W_{i,j}:=\delta_{i,j}(1-m_i^2)$.
Therefore,
\begin{equation*}
\det(I- \mathcal{U}\mathcal{M})
=
\det(Y)
\prod_{i \in V} (1-m_i^2)
\prod_{[e]\in E}(1-u_e u_{\bar{e}})
=
\text{ R.H.S. of (\ref{eqthmmain})}
\end{equation*}
For the left equality, Theorem \ref{thmEdgeIharaFormula} is used.
\end{proof}

Theorem \ref{thmmain} shows that the determinant of the Hessian of
the Bethe free energy is essentially equal to $\det(I-
\mathcal{U}\mathcal{M})$, the reciprocal of the edge zeta function.
Since the matrix $\mathcal{U}\mathcal{M}$ has a direct connection
with LBP as seen in section 5, the above formula derives many consequences shown in
the rest of the paper.

\section{Application to positive definiteness conditions} \label{subsectionPDcondition}
%

The convexity of the Bethe free energy is an important issue, as it
guarantees uniqueness of the fixed point. 
Pakzad et al \cite{PAstat} 
and Heskes \cite{Huniquness} 
derive sufficient conditions of convexity and
show that the Bethe free energy is convex for trees and graphs with
one cycle. In this section, instead of such global structure, we
shall focus the local structure of the Bethe free energy as an
application of the main formula.

For given square matrix $X$,
$\spec{X} \subset \mathbb{C}$ denotes the set of eigenvalues (spectra),
and $\rho(X)$ the spectral radius of a matrix $X$,
i.e., the maximum of the modulus of the eigenvalues.

\begin{thm}
\label{thmpositive}
Let $\mathcal{M}$ be the matrix given by (\ref{DirectedEdgeMatrix}).
For given $\{m_i,\chi_{ij}\} \in L(G)$,
$\mathcal{U}$ is defined by (\ref{defuij}).
Then, 
$ \hspace{4mm}
\spec{\mathcal{U}\mathcal{M}} \hspace{0.5mm} \subset
\hspace{0.5mm} \mathbb{C} \setminus \mathbb{R}_{\geq 1}
\quad \Longrightarrow \quad
\nabla^2 F
\text{ is a positive definite matrix at } \{m_i,\chi_{ij}\}. \label{positivitycondition}
$
\end{thm}
\vspace*{-1mm}
\begin{proof}
We define
$m_{i}(t):=m_i$ and $\chi_{ij}(t):=t \chi_{ij}+(1-t)m_i m_j$.
Then $\{m_i(t),\chi_{ij}(t)\} \in  L(G)$
and $\{m_i(1),\chi_{ij}(1)\}=\{m_i,\chi_{ij}\}$.
For $t \in [0,1]$, we define $\mathcal{U}(t)$ and $\nabla^2 F(t)$ 
in the same way by $\{m_i(t),\chi_{ij}(t)\}$.
We see that $\mathcal{U}(t)=t \mathcal{U}$.
Since 
$\spec{\mathcal{U}\mathcal{M}} \subset  \mathbb{C} \setminus \mathbb{R}_{\geq 1}$,
we have
$
\det(I- t\mathcal{U}\mathcal{M})
\neq 0$
${}^{\forall} t \in [0,1]$.
From Theorem \ref{thmmain},
$\det(\nabla^2 F(t)) \neq 0$ holds on this interval.
Using (\ref{eqYformula}) and $\chi_{ij}(0)=m_i(0)m_j(0)$,
we can check that $\nabla^2 F(0)$
is positive definite.
Since the eigenvalues of 
$\nabla^2 F(t)$ are real
and continuous with respect $t$,
the eigenvalues of $\nabla^2 F(1)$ must be positive reals.
\end{proof}
%
We define the symmetrization of  $u_{\ed{i}{j}}$ and
$u_{\ed{j}{i}}$ by
\vspace{-1mm}
\begin{equation}
\beta_{\ed{i}{j}}=\beta_{\ed{j}{i}}
:=
\frac{\chi_{ij}-m_i m_j}{\{(1-m_i^2)(1-m_j^2)\}^{1/2}}
=
\frac{ \covariance{b_{ij}}{x_i}{x_j} }
{\{\variance{b_i}{x_i} \variance{b_j}{x_j} \}^{1/2}}. \label{defbeta}
\end{equation}
Thus, $u_{\ed{i}{j}}u_{\ed{j}{i}}=\beta_{\ed{i}{j}}\beta_{\ed{j}{i}}$.
Since $\beta_{\ed{i}{j}}=\beta_{\ed{j}{i}}$,
we sometimes abbreviate $\beta_{\ed{i}{j}}$ as $\beta_{ij}$.
From the final expression, we see that $|\beta_{ij}| < 1$.
Define diagonal matrices $\mathcal{Z}$ and $\mathcal{B}$ by
$(\mathcal{Z})_{e,e'}:=\delta_{e,e'} {(1-m_{t(e)}^2)^{1/2}}$ and
$(\mathcal{B})_{e,e'}:=\delta_{e,e'} \beta_{e}$ respectively. Then
we have
$\mathcal{Z}^{}\mathcal{U}\mathcal{M}\mathcal{Z}^{-1}=\mathcal{B}\mathcal{M}$,
because
\begin{equation}
(\mathcal{Z}^{}\mathcal{U}\mathcal{M}\mathcal{Z}^{-1})_{e,e'}
=
(1-m_{t(e)}^2)^{1/2} u_{e}
(\mathcal{M})_{e,e'}
(1-m_{o(e)}^2 )^{-1/2} \\
= \beta_{e} (\mathcal{M})_{e,e'}. \nonumber
\end{equation}
Therefore $\spec{\mathcal{U}\mathcal{M}}=\spec{\mathcal{B}\mathcal{M}}$.

The following corollary gives a more explicit condition of the
region where the Hessian is positive definite in terms of the
correlation coefficients of the pseudomarginals.
%
\begin{cor} \label{corpositivedefiniteregion}
Let $\alpha$ be the Perron Frobenius eigenvalue of $\mathcal{M}$ and
define $L_{\alpha^{-1}}(G):=\{ \{m_i,\chi_{ij} \} \in L(G)|
|\beta_{e}| < \alpha^{-1} \text{ for all } e \in \vec{E}  \}$. Then, the
Hessian $\nabla^2 F$ is positive definite on $L_{\alpha^{-1}}(G)$.
\end{cor}
\vspace*{-1mm}
\begin{proof}
Since $|\beta_{e}| < \alpha^{-1}$, we have
$\rho(\mathcal{B} \mathcal{M}) < \rho(\alpha^{-1} \mathcal{M}) =1$
(\cite{HJmatrix} Theorem 8.1.18).
Therefore $\spec{\mathcal{B}\mathcal{M}} \cap \mathbb{R}_{ \geq 1}= \phi$.
\end{proof}
\vspace*{-1mm}
%
As is seen from (\ref{IharaFormula}), $\alpha^{-1}$ is the distance
from the origin to the nearest pole of Ihara's zeta $\zeta_{G}(u)$.
From example \ref{example1}, we see that $\zeta_{G}(u)=1$ for a tree
$G$ and $\zeta_{C_N}(u)=(1-u^{N})^{-2}$ for a 1-cycle graph $C_N$.
Therefore $\alpha^{-1}$ is $\infty$ and $1$ respectively. In these
cases, $L_{\alpha^{-1}}(G)=L(G)$ and $F$ is a strictly convex
function on $L(G)$, because $|\beta_{e}|<1$ always holds. This
reproduces the results shown in \cite{PAstat}. In general, using
Theorem 8.1.22 of \cite{HJmatrix}, we have 
$ \min_{i \in V} d_i -1 \leq \alpha \leq   \max_{i \in V} d_i -1. $
%

Theorem \ref{thmmain} is also useful to show non-convexity.
\begin{cor}\label{nonconvex}
Let $\{m_i(t):=0,\chi_{ij}(t):=t \}\in L(G)$ for  $ t <1$.
Then we have
\begin{equation}
\lim_{t \rightarrow 1}
\det(\nabla^2 F(t))(1-t)^{M+N-1}
=
-2^{-M-N+1}
(M-N)
\kappa(G), \label{DetformulaWithHashimoto}
\end{equation}
where $\kappa(G)$ is the number of spanning trees in $G$.
In particular,
$F$ is never convex on $L(G)$ for any connected graph with
at least two linearly independent cycles, i.e.  $M-N \geq 1$.
\end{cor}
\begin{proof}
The equation (\ref{DetformulaWithHashimoto}) is obtained by 
Hashimoto's theorem \cite{Hzeta}, which gives the
$u \rightarrow 1$ limit of the Ihara zeta function.
(See supplementary material for the detail.)
If $M-N \geq 1$, the right hand side of (\ref{DetformulaWithHashimoto}) is negative.
As approaches to $\{m_i=0,\chi_{ij}=1\} \in L(G)$, the determinant
of the Hessian diverges to $- \infty$.
Therefore the Hessian is not positive definite near the point.
\end{proof}

Summarizing the results in this section,
we conclude that $F$ is convex on $L(G)$
if and only if $G$ is a tree or a graph with one cycle.
To the best of our knowledge,
this is the first proof of this fact.

%
%
\section{Application to stability analysis} \label{secstability}
In this section we discuss the local stability of LBP and the local
structure of the Bethe free energy around a LBP fixed point.
Heskes \cite{Hstable} shows that a locally stable fixed point of
sufficiently damped LBP is a local minima of the Bethe free energy.  
The converse is not necessarily true in general, and we will elucidate
the gap between these two properties.

First, we regard the LBP update as a dynamical system.
Since the model is binary,
each message $\mu_{\ed{i}{j}}(x_j)$ is parametrized by one
parameter, say $\eta_{\ed{i}{j}}$.
The state of LBP algorithm is expressed by
$\boldsymbol{\eta}=(\eta_{e})_{e \in \vec{E}} \in C(\vec{E})$, 
and the update rule (\ref{BPupdate}) is
identified with a transform $T$ on
$C(\vec{E})$,
$\boldsymbol{\eta}^{{\rm new}} = T(\boldsymbol{\eta})$.
Then, the set of fixed points
of LBP is
$\{ \boldsymbol{\eta}^{\infty}\in C(\vec{E})| T (\boldsymbol{\eta}^{\infty}) =\boldsymbol{\eta}^{\infty}\}$.

A fixed point $\boldsymbol{\eta}^{\infty}$ is called
{\it locally stable}
if LBP starting with a point sufficiently close to
$\boldsymbol{\eta}^{\infty}$
converges to $\boldsymbol{\eta}^{\infty}$.
The local stability is determined by the linearizion $T'$
around the fixed point.
As is discussed in \cite{MKproperty},
$\boldsymbol{\eta}^{\infty}$ is locally stable
if and only if
$\spec{T'(\boldsymbol{\eta}^{\infty})}  \subset
\{ \lambda \in \mathbb{C}| |\lambda| < 1\}$.

To suppress oscillatory behaviors of LBP,
damping of update $T_{\epsilon }:=(1- \epsilon ) T+ \epsilon I$
is sometimes useful,
where $0 \leq \epsilon < 1$ is a damping strength
and $I$ is the identity.
A fixed point is {locally stable with some damping}
if and only if
$\spec{T'(\boldsymbol{\eta}^{\infty})}  \subset
\{ \lambda \in \mathbb{C}| {\rm  Re}\lambda < 1\}$.

There are many 
representations of the linearization (derivative) of LBP update
(see \cite{MKproperty,ITAinfo}), 
we choose a good coordinate
following Furtlehner et al \cite{FLFtraffic}.
In section 4 of \cite{FLFtraffic},
they transform messages as
$\mu_{\ed{i}{j}} \rightarrow \mu_{\ed{i}{j}}/ \mu^{\infty}_{\ed{i}{j}}$
and functions as
$\psi_{ij} \rightarrow b_{ij}/ (b_i b_j)$
and
$\psi_{i} \rightarrow b_{i}$,
where $\mu^{\infty}_{\ed{i}{j}}$ is the message of the fixed point.
This changes only the representations of messages and functions,
and does not affect LBP essentially.
This transformation causes
$T'(\boldsymbol{\eta}^{\infty}) \rightarrow P^{} T'(\boldsymbol{\eta}^{\infty}) P^{-1}$
with an invertible matrix $P$.
Using this transformation, we see that the following fact holds.
(See supplementary material for the detail.)
\begin{thm}
[\cite{FLFtraffic}, Proposition 4.5]
Let $u_{\ed{i}{j}}$ be given by
(\ref{DefBelief}), (\ref{defbbymchi}) and (\ref{defuij})
at a LBP fixed point $\boldsymbol{\eta}^{\infty}$.
The derivative $T'(\boldsymbol{\eta}^{\infty})$
is similar to
$\mathcal{U}\mathcal{M}$, i.e.
$\mathcal{U}\mathcal{M}=P^{} T'(\boldsymbol{\eta}^{\infty}) P^{-1}$
with an invertible matrix $P$.
\end{thm}

Since
$\det(I -T^{'}(\boldsymbol{\eta}^{\infty})) = \det(I - \mathcal{U}\mathcal{M})$,
the formula in Theorem \ref{thmmain}
implies a direct link between the linearization
$T^{'}(\boldsymbol{\eta}^{\infty})$ and
the local structure of the Bethe free energy.
From Theorem \ref{thmpositive}, we have that
a fixed point of LBP is a local minimum of the Bethe
free energy if
\hspace{0.5mm}
$\spec{T'(\boldsymbol{\eta}^{\infty})}  \subset \mathbb{C} \setminus \mathbb{R}_{\geq 1}$.
%
%

It is now clear that the condition for
positive definiteness, local stability of damped LBP and local stability
of undamped LBP are given in terms of the set of eigenvalues,
$\mathbb{C} \setminus \mathbb{R}_{\geq 1}$,
$\{ \lambda \in \mathbb{C}| {\rm  Re}\lambda < 1\}$ and
$\{ \lambda \in \mathbb{C}| |\lambda| < 1\}$ respectively.
A locally stable fixed point of sufficiently damped LBP
is a local minimum of the Bethe free energy,
because $\{ \lambda \in \mathbb{C}| {\rm  Re}\lambda < 1\}$
is included in
$\mathbb{C} \setminus \mathbb{R}_{\geq 1}$.  This reproduces Heskes's result \cite{Hstable}.
Moreover, we see the gap between the locally stable fixed points with some damping and the
local minima of the Bethe free energy:
if $\spec{T'(\boldsymbol{\eta}^{\infty})}$ is included in
$\mathbb{C} \setminus \mathbb{R}_{\geq 1}$ but not in
$\{ \lambda \in \mathbb{C}| {\rm  Re}\lambda < 1\}$,
the fixed point is a local minimum of the Bethe free energy though
it is not a locally stable fixed point of LBP with any damping.



It is interesting to ask under which condition a local minimum of the
Bethe free energy is a stable fixed point of (damped) LBP.  While we do
not know a complete answer, for an attractive model, which is defined by
$J_{ij}\geq 0$, the following theorem implies that if a stable fixed
point becomes unstable by changing $J_{ij}$ and $h_i$, the
corresponding local minimum also disappears.


\begin{thm}
\label{thmattractive}
Let us consider
continuously parametrized attractive models
$\{\psi_{ij}(t),\psi_{i}(t)\}$,
e.g. $t$ is a temperature: $\psi_{ij}(t)=\exp(t^{-1} J_{ij}x_i x_j)$
and
$\psi_{i}(t)=\exp(t^{-1} h_i x_i )$.
For given $t$, run LBP algorithm and find a (stable) fixed point.
If we continuously change $t$ and
see the LBP fixed point becomes unstable across $t=t_0$,
then the corresponding local minimum of the Bethe free energy becomes
a saddle point across $t=t_0$.
\end{thm}
\vspace{-4mm}
\begin{proof}
From (\ref{DefBelief}),
we see
$b_{ij}(x_i,x_j) \propto \exp (J_{ij}x_i x_j + \theta_i x_i+ \theta_j x_j)$
for some $\theta_i$ and $\theta_j$.
From $J_{ij} \geq 0$,
we have $\covariance{b_{ij}}{x_i}{x_j}=\chi_{ij}-m_i m_j \geq 0$, and thus $u_{\ed{i}{j}} \geq 0$.
When the LBP fixed point becomes unstable,
the Perron Frobenius eigenvalue of $\mathcal{U} \mathcal{M}$ goes over $1$, which means
$\det(I-\mathcal{U} \mathcal{M})$ crosses $0$.
From Theorem \ref{thmmain} we see that
$\det(\nabla^{2} F)$ becomes positive to negative at $t=t_0$.
\end{proof}
Theorem \ref{thmattractive} extends Theorem 2 of \cite{MKproperty}, which discusses only
the case of vanishing local fields $h_i=0$
and the trivial fixed point (i.e. $m_i=0$).


%
%
%
\section{Application to uniqueness of LBP fixed point}

The uniqueness of LBP fixed point 
is a concern of many studies, because 
the property guarantees that LBP finds the global
minimum of the Bethe free energy if it converges. 
The major approaches to the uniqueness is to consider
equivalent minimax problem \cite{Huniquness}, contraction
property of LBP dynamics \cite{IFW,MKsufficient}, and to use the theory
of Gibbs measure \cite{TJgibbsmeasure}. We will propose a different,
differential topological approach to this problem.

In our approach, in combination with Theorem \ref{thmmain},
the following theorem is the basic apparatus.

\begin{thm}
\label{thmindexsum}
If $ \det \nabla^2 F(q) \neq 0$ for all $q \in (\nabla F)^{-1}(0)$ then
\begin{equation}
\sum_{q: \nabla F(q)=0 }
\sgn
\left(
 \det \nabla^2 F(q)
\right)
=
1,
\quad
\text{ where }
\sgn(x):=
\begin{cases}
1  \quad \text{ if } x>0, \\
-1  \text{ if } x<0.
\end{cases}  \nonumber
\end{equation}
We call each summand, which is $+1$ or $-1$, the {\it index} of $F$ at $q$.
\end{thm}
Note that the set $(\nabla F)^{-1}(0)$, which is the stationary points of the
Bethe free energy, coincides with the fixed points of LBP.
The above theorem asserts that the sum of indexes
of all the fixed points must be one.
As a consequence, the number of the fixed points of LBP is always odd.
Note also that the index is a local quantity, while the assertion
expresses the global structure of the function $F$.

For the proof of Theorem \ref{thmindexsum},
we prepare two lemmas.
The proof of Lemma \ref{lemgradBethe} is shown in
the supplementary material.
Lemma \ref{lemdegreeofmap} is a standard result in differential
topology, and we refer \cite{DFN} Theorem 13.1.2 and comments in p.104
for the proof.
\begin{lem}
\label{lemgradBethe}
If a sequence $\{q_n \} \subset L(G)$ converges to a point
$q_{*} \in \partial L(G)$, then
$\lVert \nabla F(q_n) \rVert \rightarrow \infty$,
where $\partial L(G)$ is the boundary of $L(G) \subset \mathbb{R}^{N+M}$.
\end{lem}
\begin{lem}
\label{lemdegreeofmap}
Let $M_1$ and $M_2$ be compact, connected and orientable manifolds
with boundaries.
Assume that the dimensions of $M_1$ and $M_2$ are the same.
Let $f:M_1 \rightarrow M_2 $ be a smooth map satisfying
$f(\partial M_1) \subset \partial M_2 $.
For a regular value of $p \in M_2$, i.e.
$\det(\nabla f(q))\neq 0$ for all $q \in f^{-1}(p)$,
we define the degree of the map $f$ by
$
\deg f := \sum_{q \in f^{-1}(p)} \sgn( \det \nabla f(q)).
$
Then $\deg f$ does not depend on the choice of a regular value
$p \in M_2$. 
\end{lem}
%
\vspace{-2mm}
\begin{proof}[Sketch of proof]
Define a map $\Phi:L(G) \rightarrow \mathbb{R}^{N+M}$
by $\Phi:= \nabla F +  \binom{\boldsymbol{h}}{\boldsymbol{J}} $.
Note that $\Phi$ does not depend on $\boldsymbol{h}$
and $\boldsymbol{J}$ as seen from (\ref{BetheFreeEnergybinary}).
Then it is enough to prove
\begin{equation}
\sum_{q \in \Phi^{-1}
(\binom{\boldsymbol{h}}{\boldsymbol{J}})
}
\sgn( \det \nabla \Phi(q))
=
\sum_{q \in \Phi^{-1}
(0)
}
\sgn( \det \nabla \Phi(q)),
\label{eq1thmindexsum}
\end{equation}
because $\Phi^{-1}(0)$ has a unique element
$\{m_i=0,\chi_{ij}=0\}$, at which $\nabla^{2}F$ is positive definite,  
and the right hand side of (\ref{eq1thmindexsum})
is equal to one. 
Define a sequence of manifolds $\{C_n\}$ by 
$C_n:=\{q\in L(G)|\sum_{ij \in E}\sum_{x_i,x_j} \hspace{-2mm}-\log b_{ij} \leq n \}$,
which increasingly converges to $L(G)$.
Take $K>0$ and $\epsilon >0$ to satisfy
$K -\epsilon > \lVert \binom{\boldsymbol{h}}{\boldsymbol{J}}  \rVert$.
From Lemma \ref{lemgradBethe}, for sufficiently large $n_0$, 
we have 
$\Phi^{-1}(0),\Phi^{-1} \binom{\boldsymbol{h}}{\boldsymbol{J}} \subset
 C_{n_0}$ and
$\Phi(\partial C_{n_0}) \cap B_0(K)=\phi$,
where $B_0(K)$ is the closed ball of radius $K$ at the origin.
Let $\Pi_{\epsilon}:\mathbb{R}^{N+M} \rightarrow  B_0(K)$ be a smooth
 map that is the identity on $B_0(K-\epsilon)$,
monotonically increasing on $\lVert x \rVert$, and
 $\Pi_{\epsilon}(x)=\frac{K}{\lVert x \rVert}x$
for $\lVert x \rVert \geq K$.
We obtain a map
$\tilde{\Phi }:=\Pi_{\epsilon} \circ \Phi :C_{n_0} \rightarrow  B_0(K)$
such that $\tilde{\Phi }(\partial C_{n_0}) \subset \partial B_0(K)$.
Applying Lemma \ref{lemdegreeofmap} yields (\ref{eq1thmindexsum}).
\end{proof}

If we can guarantee that the index
of every fixed point is $+1$
in advance of running LBP,
we conclude that fixed point of LBP is unique.
We have the following a priori information for $\beta$.
%
\begin{lem}
Let $\beta_{ij}$ be given by (\ref{defbeta})
at any fixed point of LBP.
Then $|\beta_{ij}| \leq \tanh(|J_{ij}|)$ and
$\sgn(\beta_{ij})=\sgn(J_{ij})$ hold.
\end{lem}
\vspace*{-2mm}
\begin{proof}
From (\ref{DefBelief}),
we see that
$b_{ij}(x_i,x_j) \propto \exp (J_{ij}x_i x_j + \theta_i x_i+ \theta_j x_j)$
for some $\theta_i$ and $\theta_j$.
With (\ref{defbeta}) and straightforward computation,
we obtain
$\beta_{ij}= \sinh(2 J_{ij})
(\cosh(2 \theta_i)+\cosh(2 J_{ij}))^{-1/2}
(\cosh(2 \theta_j)+\cosh(2 J_{ij}))^{-1/2}$.
The bound is attained when $\theta_{i}=0$ and $\theta_{j}=0$.
\end{proof}

%
From Theorem \ref{thmindexsum} and Lemma \ref{betabound}, we can
immediately obtain the uniqueness condition in \cite{MKsufficient},
though the stronger contractive property is proved under the same condition in \cite{MKsufficient}.

\begin{figure}
\begin{minipage}{.35\linewidth}
\begin{center}
\includegraphics[scale=0.28]{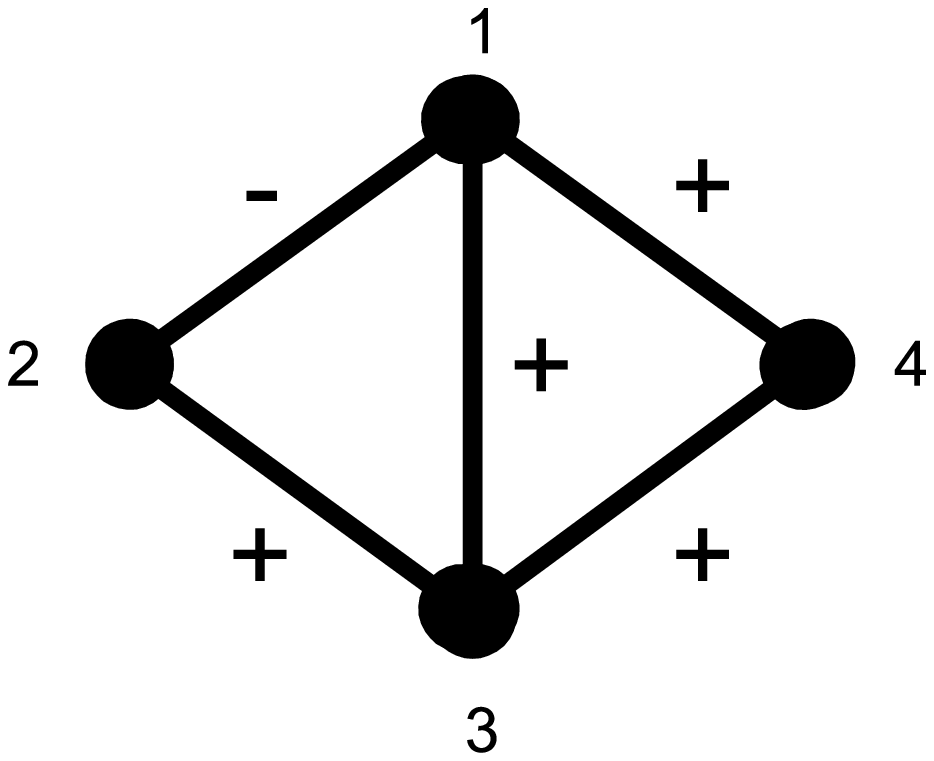}
\vspace{-2mm}
\caption{Graph of Example \ref{example2}. \label{figureExampleGraph1}}
\end{center}
\end{minipage}
\begin{minipage}{.3\linewidth}
\begin{center}
\includegraphics[scale=0.26]{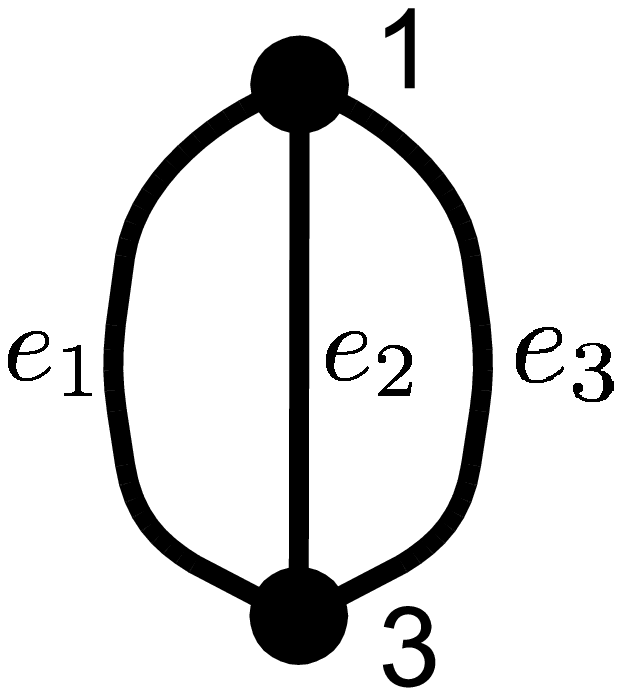}
\vspace{-1mm}
\caption{Graph $\cg{G}$. \label{figureExampleGraph2}}
\end{center}
\end{minipage}
\begin{minipage}{.3\linewidth}
\begin{center}
\includegraphics[scale=0.35]{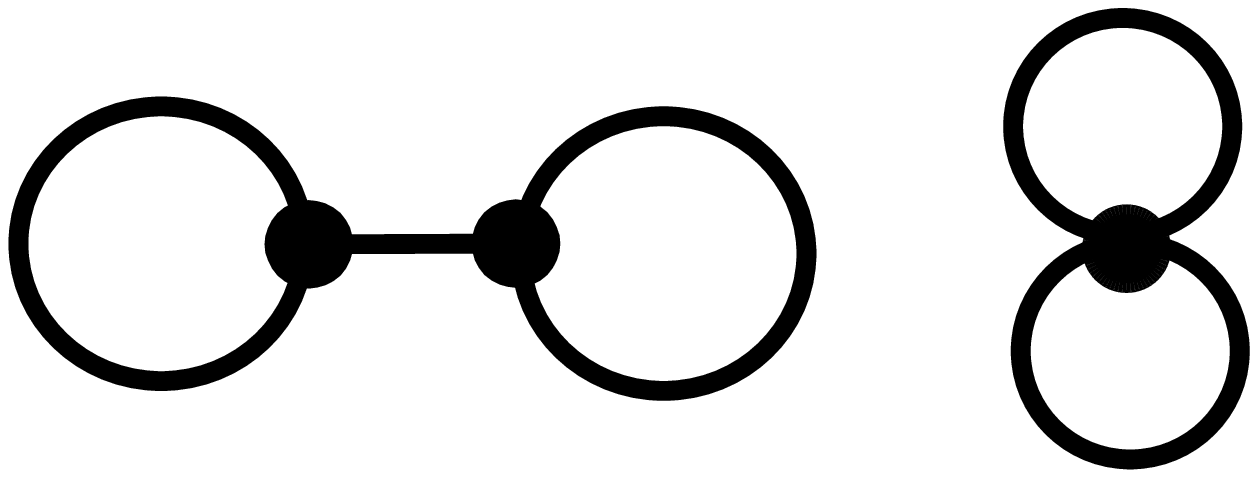}
\caption{Two other types. \label{figureExampleGraph3}}
\end{center}
\end{minipage}
\vspace{-3mm}
\end{figure}

\begin{cor}
[\cite{MKsufficient}] \label{betabound}
If $\rho(\mathcal{J}\mathcal{M}) < 1$,
then the fixed point of LBP is unique,
where $\mathcal{J}$ is a diagonal matrix defined by
$\mathcal{J}_{e,e'}=\tanh(|J_{e}|) \delta_{e,e'}$.
\end{cor}
\vspace*{-2mm}
\begin{proof}
Since $|\beta_{ij}| \leq \tanh(|J_{ij}|)$,
we have
$\rho(\mathcal{B}\mathcal{M}) \leq \rho(\mathcal{J}\mathcal{M}) <1$.
(\cite{HJmatrix} Theorem 8.1.18.)
Then
$\det(I - \mathcal{B}\mathcal{M})=\det(I -\mathcal{U}\mathcal{M})>0$
implies that the index of any LBP
 fixed point must be $+1$.
\end{proof}
%
%
%
In the proof of the above corollary,
we only used the bound of modulus.
In the following case of Corollary \ref{cor2loop},
we can utilize
the information of signs.
To state the corollary, we need a terminology.
The interactions $\{J_{ij},h_i\}$ and $\{J'_{ij},h'_i\}$ are
said to be {\it equivalent} if
there exists $(s_i) \in \{\pm 1\}^{V}$ such that $J'_{ij}=J_{ij}s_i s_j$
and $h'_i=h_i s_i$.
Since an equivalent model is obtained by gauge transformation
$x_i \rightarrow x_i s_i$, the uniqueness property of LBP for equivalent models is unchanged.
\begin{cor} 
\label{cor2loop}
If the number of linearly independent cycle of $G$ is two
(i.e. $M-N+1=2$),
and the interaction is not equivalent to attractive model,
then the LBP fixed point is unique.
\end{cor}
%
The proof is shown in the supplementary material.
We give an example to illustrate the outline.
\begin{example}
\label{example2}
Let $V:=\{1,2,3,4\}$ and $E:=\{12,13,14,23,34\}$.
The interactions are given by
arbitrary $\{h_i\}$ and
$\{-J_{12},J_{13},J_{14},J_{23},J_{34}\}$
with $J_{ij} \geq 0$. See figure \ref{figureExampleGraph1}.
It is enough to check that
$\det(I -  \mathcal{B}\mathcal{M}) > 0$
for arbitrary
$0 \leq \beta_{13},\beta_{23},\beta_{14},\beta_{34} < 1$
and $-1 < \beta_{12} \leq 0$.
Since the prime cycles of $G$ bijectively correspond
to those of $\cg{G}$ (in figure \ref{figureExampleGraph2}),
we have
$\det(I -  \mathcal{B}\mathcal{M}) = \det(I - \mathcal{\cg{B}}\mathcal{\cg{M}})$, 
where $\cg{\beta}_{e_1}=\beta_{12}\beta_{23}$,
$\cg{\beta}_{e_2}=\beta_{13}$,
and $\cg{\beta}_{e_3}=\beta_{34}$.
We see that
$\det(I -  \mathcal{\cg{B}}\mathcal{\cg{M}})=
(1 - \cg{\beta}_{e_1} \cg{\beta}_{e_2} - \cg{\beta}_{e_1}
 \cg{\beta}_{e_3}  - \cg{\beta}_{e_2} \cg{\beta}_{e_3}
- 2 \cg{\beta}_{e_1} \cg{\beta}_{e_2} \cg{\beta}_{e_3} )
(1 -\cg{\beta}_{e_1} \cg{\beta}_{e_2} -\cg{\beta}_{e_1} \cg{\beta}_{e_3}
- \cg{\beta}_{e_2} \cg{\beta}_{e_3}
+ 2 \cg{\beta}_{e_1} \cg{\beta}_{e_2} \cg{\beta}_{e_3} ) > 0$.
In other cases, we can reduce to the graph $\cg{G}$ or the graphs in
figure \ref{figureExampleGraph3} similarly (see the supplementary material).
\end{example}

For attractive models, the fixed point of the LBP is not necessarily
unique.


For graphs with multiple cycles, all the existing results on uniqueness make 
assumptions that upperbound $|J_{ij}|$ essentially. In contrast,
Corollary \ref{cor2loop} applies to arbitrary strength of interactions if
the graph has two cycles and the interactions are not attractive. 
It is noteworthy that, from Corollary \ref{nonconvex},
the Bethe free energy is non-convex in the situation of Corollary
\ref{cor2loop}, while the fixed point is unique.

\section{Concluding remarks}
For binary pairwise models,
we show the connection between the edge zeta function and
the Bethe free energy in Theorem \ref{thmmain}, in the proof of which
the multi-variable version of Ihara's formula (Theorem
\ref{thmEdgeIharaFormula}) is essential.
After the initial submission of this paper,
we found that Theorem \ref{thmmain} is extended to
a more general class of models 
including multinomial models and Gaussian models
represented by arbitrary factor graphs. 
We will discuss the extended formula and its applications
in a future paper.
%

Some recent researches on LBP have suggested 
the importance of zeta function.
In the context of the LDPC code,
which is an important application of LBP,
Koetter et al \cite{KLVWpseudo,KLVWcharacterizations}
show the connection between pseudo-codewords and the edge zeta function.
On the LBP for the Gaussian graphical model,
Johnson et al \cite{JCC} give
zeta-like product formula of the partition function.
While these are not directly related to our work,
pursuing covered connections is an interesting future research topic.

\subsubsection*{Acknowledgements}
This work was supported in part by Grant-in-Aid for JSPS Fellows
20-993 and Grant-in-Aid for Scientific Research (C) 19500249.

\subsubsection*{References}
\bibliographystyle{unsrt}
\bibliography{NIPS2009}

\newpage

\input{supplement.tex}

\end{document}

%% file: supplement.tex
\part*{
Supplementary Material for \\
\vspace{3mm}
{\large ``Graph Zeta Function in the Bethe Free Energy and \\ 
Loopy Belief Propagation''  \\
Yusuke Watanabe and Kenji Fukumizu, NIPS 2009 }
}
\hrule

\appendix
\vspace{5mm}
\section{Proof of Theorem \ref{thmEdgeIharaFormula}}
\begin{proof}
First, we define three linear operators 
$\mathcal{O}:C(V) \rightarrow C(\vec{E})$,
$\mathcal{T}^*:C(\vec{E})\rightarrow C(V)$,
and
$\iota:C(\vec{E})\rightarrow C(\vec{E})$ as follows:
\begin{equation}
(\mathcal{O}f)(e):=f(o(e)),
\quad
(\mathcal{T}^*g)(i):=
\hspace{-4mm}
\sum_{e \in \vec{E}, t(e)=i}
\hspace{-3mm}
g(e),
\quad
(\iota g)(e):=g(\bar{e})
\quad
\text{  where } f \in C(V)
\text{  and } g \in C(\vec{E}). \nonumber
\end{equation}

We see that
$\mathcal{M}=\mathcal{O}\mathcal{T}^*-\iota$,
because 
\begin{align*}
\big(
(\mathcal{O}\mathcal{T}^*-\iota)g
\big)
(e)
=
\hspace{-2mm}
\sum_{e' \in \vec{E}, t(e')=o(e)}
\hspace{-6mm}g(e')
\hspace{2mm}
-g(\bar{e}) 
= (\mathcal{M}g)(e)
\quad
\text{ for } 
g \in C(\vec{E}).
\end{align*}

Then we have
\begin{align*}
\det(I- \mathcal{U}\mathcal{M}) 
&=
\det
\Big(
I- \mathcal{U}\mathcal{O}\mathcal{T}^*(I+\mathcal{U}\iota)^{-1}
\Big)
\det(I+\mathcal{U}\iota)  \\
&=
\det
\Big(
I-\mathcal{T}^*(I+\mathcal{U}\iota)^{-1} \mathcal{U}\mathcal{O}
\Big)
\det(I+\mathcal{U}\iota). 
\end{align*}
In the second equality, we used $\det(I_n-AB)=\det(I_m-BA)$
for $n \times m $ and $m \times n$ matrices $A$ and $B$ 
(\cite{Godsil}, Lemma 8.2.4).
The linear operator $\iota$ is 
a block diagonal matrix with standard basis.
The
$(e,\bar{e})$ block of $I+\mathcal{U}\iota$ is 
\begin{equation*}
 \begin{bmatrix}
1 & u_e \\
u_{\bar{e}} & 1
\end{bmatrix}. \nonumber
\end{equation*}
Therefore, we have
$\det(I+\mathcal{U}\iota)=\prod_{[e]\in E}(1-u_e u_{\bar{e}})$.

Finally, we check that
$\mathcal{T}^*(I+\mathcal{U}\iota)^{-1} \mathcal{U}\mathcal{O}=
\mathcal{\ev{A}}-\mathcal{\ev{D}}$.
The matrix $(I+\mathcal{U}\iota)^{-1}$ is a block diagonal matrix
with  $(e,\bar{e})$ block
\begin{equation}
\frac{1}{1-u_{e}u_{\bar{e}}}
\begin{bmatrix}
1 & -u_e \\
-u_{\bar{e}} & 1
\end{bmatrix}. 
\end{equation}
For $f \in C(V)$,
we have
\begin{align*}
\Big(
\mathcal{T}^*(I+\mathcal{U}\iota)^{-1} \mathcal{U}\mathcal{O}f
\Big)(i)
&=
\sum_{e \in \vec{E}, t(e)=i}
\Big(
(I+\mathcal{U}\iota)^{-1} \mathcal{U}\mathcal{O}f
\Big)(e) \nonumber \\
&=
\sum_{e \in \vec{E}, t(e)=i}
\frac{1}{1-u_{e}u_{\bar{e}}}
\Big(
(
\mathcal{U}\mathcal{O}f
)(e)
-
u_{e}
(
\mathcal{U}\mathcal{O}f
)(\bar{e})
\Big) \nonumber \\
&=
\sum_{e \in \vec{E}, t(e)=i}
\frac{1}{1-u_{e}u_{\bar{e}}}
\Big(
u_{e}f(o(e))
-
u_{e}u_{\bar{e}}
f(o(\bar{e}))
\Big) \nonumber \\
&=
(\mathcal{\ev{A}}f)(i)-(\mathcal{\ev{D}}f)(i).
\end{align*}
\end{proof}

\section{Proof of Theorem \ref{thmmain}}
\subsection{Explicit formula of derivatives of the Bethe free energy}
In the proof of Theorem \ref{thmmain},
the graph $G=(V,E)$ is assumed to be a simple graph, i.e.,
there is no multiple edges and loop-edge

For the proof,
we need explicit expressions of the second derivatives of the Bethe free
energy. We list them below.

The first derivatives of the Bethe Free Energy are
\begin{align}
\pd{F}{m_i}& = -h_i +
(1-d_i) \frac{1}{2} \sum_{x_i = \pm 1} x_i \log b_i(x_i)
+ \frac{1}{4}\sum_{k \in N_i}\sum_{x_i,x_k=\pm 1}x_i \log
 b_{ik}(x_i,x_k), \label{Bethem} \\
\pd{F}{\chi_{ij}}& = -J_{ij} +
\frac{1}{4} \sum_{x_i,x_j=\pm 1}x_i x_j \log  b_{ij}(x_i,x_j). \label{Bethechi}
\end{align}
%
The second derivatives of the Bethe Free Energy are
\begin{align}
\pds{F}{m_i}{m_j}
&=
\begin{cases}
(1-d_i)\frac{1}{1-m_i^2}+
\frac{1}{4}\sum_{k \in N_i}\sum_{x_i,x_k}
\frac{1}{1+m_i x_i+m_k x_k +\chi_{ik}x_i x_k }
\quad \text{ if } i=j, \\
\frac{1}{4}\sum_{x_i,x_j}
\frac{x_i x_j }{1+m_i x_i+m_j x_j +\chi_{ij}x_i x_j }
\quad \text{ if }i \text{ and } j \text{ are adjacent } (i \neq j), \\
0 
\hspace{47mm} \text{ otherwise,}
\end{cases} \\
\pds{F}{m_k}{\chi_{ij}}
&=
\begin{cases}
\frac{1}{4}
\sum_{x_i,x_j}\frac{x_j}{1+m_i x_i+m_j x_j +\chi_{ij}x_i x_j } 
\quad \text{ if } k=i, \\
\frac{1}{4}
\sum_{x_i,x_j}\frac{x_i}{1+m_i x_i+m_j x_j +\chi_{ij}x_i x_j } 
\quad \text{ if } k=j, \\
0
\hspace{47mm} \text{ otherwise, } \\
\end{cases} \\
\pds{F}{\chi_{ij}}{\chi_{kl}}
&=
\begin{cases}
\frac{1}{4}
\sum_{x_i x_j} \frac{1}{1+m_i x_i+m_j x_j +\chi_{ij}x_i x_j } 
\hspace{1mm}
\quad \text{ if } ij=kl, \\
0 
\hspace{47mm} \text{ otherwise. } \label{Bethechichi} \\
\end{cases} 
\end{align}

We use notations
\begin{align}
r_{ij}:=\frac{1}{4}\sum_{x_i,x_j}
\frac{1}{1+m_i x_i+m_j x_j +\chi_{ij}x_i x_j },\label{defr} \\
s_{ij}:=\frac{1}{4}\sum_{x_i,x_j}
\frac{x_j}{1+m_i x_i+m_j x_j +\chi_{ij}x_i x_j }, \label{defs}\\
t_{ij}:=\frac{1}{4}\sum_{x_i,x_j}
\frac{x_i x_j}{1+m_i x_i+m_j x_j +\chi_{ij}x_i x_j }. \label{deft} 
\end{align}
Note that $r_{ij}=r_{ji}$ and $t_{ij}=t_{ji}$, but 
$s_{ij}\neq s_{ji}$ in general.

\subsection{Detailed proof of Theorem \ref{thmmain}}
\begin{proof}
First, note that the Hessian of the Bethe free energy is
a square matrix of size $N+M$:
\begin{equation*}
\nabla^2 F(\{m_i,\chi_{ij}\})
:=
\begin{bmatrix}
\Big( \pds{F}{m_i}{m_j} \Big) 
& 
\Big( \pds{F}{m_i}{\chi_{st}} \Big) \\
\Big( \pds{F}{\chi_{uv}}{m_j} \Big)
& 
\Big( \pds{F}{\chi_{uv}}{\chi_{st}} \Big)
\end{bmatrix}.
\end{equation*}
Recall that $N$ is the number of
vertices and $M$ is the number of undirected edges.

\textbf{Step1}: Computation of Y \\
From 
(\ref{Bethechichi}),
the (E,E)-block of the Hessian is a diagonal matrix given by
\begin{equation*}
 \pds{F}{\chi_{ij}}{\chi_{kl}}
=
\delta_{ij,kl}r_{ij}.
\end{equation*}
Using this diagonal block, we erase (V,E)-block and 
(E,V)-block of the Hessian.
Thus, we obtain a square matrix $X$ such that $\det X =1$ and 
\begin{equation*}
X^T (\nabla^2 F) X
=
\begin{bmatrix}
\quad Y & 0 \\
\quad 0 & 
\Big( \pds{F}{\chi_{ij}}{\chi_{kl}} \Big) 
\end{bmatrix}. 
\end{equation*}
Applying an identity
\begin{small}
\begin{equation*}
\begin{pmatrix}
1& 0 & \frac{-s_{ij}}{r_{ij}} \\
0& 1 & \frac{-s_{ji}}{r_{ij}} \\
0& 0 & 1  
\end{pmatrix}
\begin{pmatrix}
w_i& t_{ij} & s_{ij} \\
t_{ij}& w_j & s_{ji} \\
s_{ij}& s_{ji} & r_{ij}  
\end{pmatrix}
\begin{pmatrix}
1& 0 & 0 \\
0& 1 & 0 \\
\frac{-s_{ij}}{r_{ij}}& \frac{-s_{ji}}{r_{ij}} & 1  
\end{pmatrix}
=
\begin{pmatrix}
w_{i}- \frac{s_{ij}^2}{r_{ij}}& t_{ij}- \frac{s_{ij}s_{ji}}{r_{ij}} & 0 \\
t_{ij}- \frac{s_{ij}s_{ji}}{r_{ij}} & w_{j} -\frac{s_{ji}^2}{r_{ij}} & 0 \\
0& 0 & r_{ij}  
\end{pmatrix} \nonumber
\end{equation*}
\end{small}
for each edge,
we have
\begin{align*}
(Y)_{i,j}
&=
\begin{cases}
(1-d_i)\frac{1}{1-m_i^2}+
\sum_{k \in N_i}
(r_{ik}-\frac{{s_{ik}^{2}}}{r_{ik}})
\quad \text{ if } i=j,  \\
t_{ij}-\frac{s_{ij}s_{ji}}{r_{ij}}
\quad \text{ if }i \text{ and } j \text{ are adjacent } , \\
0 
\quad \qquad \qquad  \text{ otherwise.}
\end{cases}
\end{align*}
The elements of $Y$ are represented in terms of $\{m_i,\chi_{ij}\}$ as follows:
\begin{align*}
(Y)_{i,i}
&=
\frac{1}{1-m_i^2}+
\sum_{k \in N_i}
(r_{ik}-\frac{{s_{ik}^{2}}}{r_{ik}}
-\frac{1}{1-m_i^2}) \nonumber \\
&=
\frac{1}{1-m_i^2}
+
\sum_{k \in N_i}
\frac{(\chi_{ik}-m_i m_k)^{2}}
{(1 - m_i^2) (1 - m_i^2 - m_k^2+ 2 m_i m_k \chi_{ik}   - \chi_{ik}^2)} 
\quad \text{ and, } \\ 
(Y)_{i,j}
&=
t_{ij}-\frac{s_{ij}s_{ji}}{r_{ij}} \nonumber \\
&=
\frac{-(\chi_{ij}-m_i m_j)}
{(1 - m_i^2 - m_j^2+ 2 m_i m_j \chi_{ij}   - \chi_{ij}^2)} 
\quad \text{ for adjacent } i \text{ and } j.
\end{align*}

\textbf{Step2}: Computation of $I_N + \mathcal{\ev{D}} - \mathcal{\ev{A}}$ \\
From the definition (\ref{defuij}) of $u_{\ed{j}{i}}$, we see that
\begin{align*}
\frac{u_{\ed{i}{j}} u_{\ed{j}{i}}}{1 - u_{\ed{i}{j}} u_{\ed{j}{i}} }
&=
\frac{(\chi_{ij}-m_i m_j)^{2}}
{ (1 - m_i^2 - m_j^2+ 2 m_i m_j \chi_{ij}   -
 \chi_{ij}^2)}, \\
\frac{u_{\ed{i}{j}}}{1 - u_{\ed{i}{j}} u_{\ed{j}{i}} }
&=
\frac{(1 - m_i^2)(\chi_{ij}-m_i m_j)}
{(1 - m_i^2 - m_j^2+ 2 m_i m_j \chi_{ij}   - \chi_{ij}^2)}.
\end{align*}

Therefore, the diagonal element is
\begin{align*}
(I_N + \mathcal{\ev{D}} - \mathcal{\ev{A}})_{i,i}
=
(I_N + \mathcal{\ev{D}})_{i,i} 
&=
1+
\sum_{k \in N_i}
\frac{u_{\ed{i}{k}} u_{\ed{k}{i}}}{1 - u_{\ed{i}{k}} u_{\ed{k}{i}} }
 \nonumber \\
&=
1
+
\sum_{k \in N_i}
\frac{(\chi_{ik}-m_i m_k)^{2}}
{(1 - m_i^2 - m_k^2+ 2 m_i m_k \chi_{ik}   - \chi_{ik}^2)},
\end{align*}
and 
for adjacent $i$ and $j$,
\begin{align*}
(I_N + \mathcal{\ev{D}} - \mathcal{\ev{A}})_{i,j}
=
- (\mathcal{\ev{A}})_{i,j}
&=
\frac{-u_{\ed{j}{i}}}{1 - u_{\ed{i}{j}} u_{\ed{j}{i}} } \nonumber \\
&=
\frac{-(1-m_j^2)(\chi_{ij}-m_i m_j)}
{(1 - m_i^2 - m_j^2+ 2 m_i m_j \chi_{ij}   - \chi_{ij}^2)}.
\end{align*}

Combining the results of step 1 and 2, we have
\begin{equation*}
I_N + \mathcal{\ev{D}} - \mathcal{\ev{A}}
=
Y
\left[
\begin{array}{cccc}
1-m_1^2 & 0 & \ldots & 0       \\
0       & 1-m_2^2 & \dots & 0  \\
\vdots & \vdots  & \ddots &  \vdots                 \\ 
0 & 0& \ldots & 1-m_N^2
\end{array}
\right].
\end{equation*}

\textbf{Step3}: Final step\\
We see that
\begin{align}
\zeta_{G}(\boldsymbol{u})^{-1} 
&=
\det(I- \mathcal{U}\mathcal{M})   \label{eqstep3a1}  \\
&=
\det(I_N + \mathcal{\ev{D}} - \mathcal{\ev{A}})
\prod_{[e]\in E}(1-u_e u_{\bar{e}})  \label{eqstep3a2}   \\
&=
\det(Y)
\prod_{i \in V} (1-m_i^2)
\prod_{[e]\in E}(1-u_e u_{\bar{e}})  \nonumber \\
&=
\det(\nabla^2 F)
\prod_{i \in V} (1-m_i^2)
\prod_{ij \in E}\frac{ 1 - u_{\ed{i}{j}} u_{\ed{j}{i}}  }{r_{ij}}
 \nonumber \\
&=
\det(\nabla^2 F)
\prod_{i \in V} (1-m_i^2)^{1-d_i}
\prod_{ij \in E}
\frac{ (1 - u_{\ed{i}{j}} u_{\ed{j}{i}})(1- m_i^2) (1 -m_j^2)  }
{r_{ij}}. \label{eqstep3a}
\end{align}
From (\ref{eqstep3a1}) to (\ref{eqstep3a2}),
we used the edge zeta version of Ihara's formula (Theorem \ref{thmEdgeIharaFormula}).

Furthermore, with a straightforward computation we see that
\begin{align*}
&\frac{ (1 - u_{\ed{i}{j}} u_{\ed{j}{i}})(1- m_i^2) (1 - m_j^2)
 }{r_{ij}}  
=
4^{4}
\prod_{x_i ,x_j= \pm 1}  b_{ij}(x_i,x_j),\\
&(1-m_i^{2})^{1-d_i}
=
2^{2-2 d_i}
\prod_{x_i = \pm 1}  b_{i}(x_i)^{1-d_i},
\end{align*}
where 
$b_{ij}(x_i,x_j)=\frac{1}{4}(1+m_i x_i + m_j x_j + \chi_{ij}x_i x_j )$
and
$b_{i}(x_i)=\frac{1}{2}(1+m_i)$.

Therefore,
\begin{align}
\text{(\ref{eqstep3a})}
&=
2^{\sum_{i \in V}(2-2 d_i) }
4^{4 M}
\det(\nabla^2 F)
\prod_{i \in V} \prod_{x_i = \pm 1}  b_{i}(x_i)^{1-d_i}
\prod_{ij \in E} \prod_{x_i ,x_j= \pm 1}  b_{ij}(x_i,x_j) \nonumber \\
&=
2^{2N+4M}
\det(\nabla^2 F)
\prod_{i \in V} \prod_{x_i = \pm 1}  b_{i}(x_i)^{1-d_i}
\prod_{ij \in E} \prod_{x_i ,x_j= \pm 1}  b_{ij}(x_i,x_j). \nonumber
\end{align}
\end{proof}

\setcounter{cor}{1}
\section{Proof of Corollary \ref{nonconvex}}
Here, we prove the limit formula in Corollary \ref{nonconvex}.

\begin{proof}
We can easily check that
$u_{\ed{i}{j}}(t)=t$, 
\begin{align*}
&\prod_{ij \in E}\prod_{x_i, x_j = \pm 1}b_{ij}(x_i,x_j) =
 4^{-4M}(1-t)^{2M}(1+t)^{2M}, \text{ and} \\
&\prod_{i \in V}\prod_{x_i = \pm 1}b_{i}(x_i)^{1-d_i}
= 2^{-2N+4M}
\end{align*}
on this interval.
Therefore
\begin{align*}
\lim_{t \rightarrow 1}
\det(\nabla^2 F(t))(1-t)^{M+N-1}
&=
\lim_{t \rightarrow 1}
\zeta_{G}(\boldsymbol{u}(t))^{-1}
(1-t)^{M+N-1} \nonumber \\
& \qquad
\Big(
  4^{-4M}(1-t)^{2M}(1+t)^{2M}
2^{-2N+4M}
2^{2N+4M}
\Big)^{-1}  \\
&=
\lim_{t \rightarrow 1}
\zeta_{G}(t)^{-1}(1-t)^{-M+N-1}
2^{-2M} \\
&=-(M-N)\kappa (G)
2^{-M-N+1}.
\end{align*} 

On the final equality,
we used Hashimoto's formula:
\begin{equation}
\lim_{u \rightarrow 1}
\zeta_{G}(u)^{-1}(1-u)^{-M+N-1}
=
-2^{M-N+1}(M-N)
\kappa(G). \nonumber
\end{equation}
We refer to \cite{Hzeta,KSzeta,Nnote} for this formula.
\end{proof}

\section{Transformation of messages and proof of Theorem 5}
\subsection{Transformation of messages}
First, we make an easy observation on the LBP update.
\begin{prop} 
\label{propchange}
Let $  \{\pi_{\ed{i}{j}}\} $ be any set of messages.
We define a transformation from
messages $ \{{\mu}^{t} _{\ed{i}{j}}\} $ to
messages $ \{\tilde{\mu}^{t} _{\ed{i}{j}}\} $ by
\begin{equation}
 \tilde{\mu}^{t}_{\ed{i}{j}}(x_j)\propto
\frac{\mu^{t}_{\ed{i}{j}}(x_j)}{\pi_{\ed{i}{j}}(x_j) } \label{propchangeeq1}.
\end{equation}
We also define transformation from functions $\{\psi_{ij},\psi_i\}$
to  functions $\{\tilde{\psi}_{ij},\tilde{\psi}_i\}$ by
\begin{align}
&\tilde{\psi}_{ij}(x_i,x_j) \propto
\frac{\psi_{ij}(x_i,x_j)}{\pi_{\ed{i}{j}}(x_j)\pi_{\ed{j}{i}}(x_i)}, \label{propchangeeq2}\\
&\tilde{\psi}_{i}(x_i) \propto
\psi_{i}(x_i) 
\prod_{k \in N_i} \pi_{\ed{k}{i}}(x_i). \label{propchangeeq3}
\end{align}
Then the update
\begin{equation}
\mu^{t+1}_{\ed{i}{j}}(x_j)
\propto
\sum_{x_i}
\psi_{ji}(x_j,x_i) \psi_{i}(x_i)
\prod_{k \in N_i \setminus j}
\mu^{t}_{\ed{k}{i}}(x_i),       \label{propchangeupdate1}
\vspace{-1mm}
\end{equation}
is equivalent to
\begin{equation}
\tilde{\mu}^{t+1}_{\ed{i}{j}}(x_j)
\propto
\sum_{x_i}
\tilde{\psi}_{ji}(x_j,x_i) \tilde{\psi}_{i}(x_i)
\prod_{k \in N_i \setminus j}
\tilde{\mu}^{t}_{\ed{k}{i}}(x_i).          \label{propchangeupdate2}
\vspace{-1mm}
\end{equation}
\end{prop}
\begin{proof}
The equivalence of (\ref{propchangeupdate1}) and (\ref{propchangeupdate2})
is easily checked by (\ref{propchangeeq1}),
 (\ref{propchangeeq2}), and  (\ref{propchangeeq3}).
\end{proof}
Symbolically, Proposition \ref{propchange} implies that 
\begin{equation}
\Pi \circ T \circ \Pi^{-1} = \tilde{T},
\end{equation}
where $\Pi$ is the transformation of the messages by
$\pi_{\ed{i}{j}}$,
$T$ is the LBP update with $\{\psi_{ij},\psi_{i}\}$, and
$\tilde{T}$ is the LBP update with
$\{\tilde{\psi}_{ij},\tilde{\psi}_{i}\}$ .
Differentiation of this relation gives the transformation
of the linearization matrix.

If we choose  $  \{\pi_{\ed{i}{j}}\}  $ as
$ \pi_{\ed{i}{j}}(x_j) =  \mu^{\infty}_{\ed{i}{j}}(x_j)$,
then (\ref{propchangeeq1}),
 (\ref{propchangeeq2}) and  (\ref{propchangeeq3}) becomes
\begin{align}
&\tilde{\mu}^{t}_{\ed{i}{j}}(x_j)\propto 
\frac{\mu^{t}_{\ed{i}{j}}(x_j)}{\mu^{\infty}_{\ed{i}{j}}(x_j) }
 \label{changefixedeq1} \\
&\tilde{\psi}_{ij}(x_i,x_j) \propto
\frac{b_{ij}(x_i,x_j)}{b_{i}(x_i)b_j(x_j)} \label{changefixedeq2}\\
&\tilde{\psi}_{i}(x_i) \propto
b_{i}(x_i).  \label{changefixedeq3}
\end{align}
This is the transformation used in the paper.

\subsection{Proof of Theorem 5}

\begin{proof}
Let $\{\mu^{\infty}_{\ed{i}{j}}(x_j)\}$
be the set of messages at the fixed point and
let $\Pi$ be the transformation of messages defined by
the fixed point messages.
We parameterize the messages by
$\eta_{\ed{i}{j}}^{t}=\mu^{t}_{\ed{i}{j}}(+) /
 \mu^{t}_{\ed{i}{j}}(-)$.
It is enough to prove the assertion 
after the transformation and
in this parameterization,
because these operations cause similar
linearization matrices.

After the transformation,
the LBP update is given in terms of 
$\tilde{\eta}$ as follows:
\begin{align*}
 \tilde{\eta}^{t+1}_{\ed{i}{j}}
&=
\frac
{ \sum_{x_i}\tilde{\psi}_{ji}(+,x_i)\tilde{\psi_{i}}(x_i) \prod_{k \in N_i \setminus j}
 \tilde{\mu}_{\ed{k}{i}}(x_i)}
{ \sum_{x_i}\tilde{\psi}_{ji}(-,x_i)\tilde{\psi_{i}}(x_i) \prod_{k \in N_i \setminus j}
 \tilde{\mu}_{\ed{k}{i}}(x_i)} \nonumber \\
&=
\frac
{ \frac{b_{ji}(+,+)}{b_{j}(+)}  \prod_{k \in N_i \setminus j}
 \tilde{\eta}_{\ed{k}{i}}(x_i)
+\frac{b_{ji}(+,-)}{b_{j}(+)}  }
{ \frac{b_{ji}(-,+)}{b_{j}(-)}  \prod_{k \in N_i \setminus j}
 \tilde{\eta}_{\ed{k}{i}}(x_i)
+\frac{b_{ji}(-,-)}{b_{j}(-)}  }.
\end{align*}

Let ${\boldsymbol{\tilde{\eta}}^{\infty}}:=\Pi(\boldsymbol{\eta}^{\infty})$,
then  $\tilde{\eta}^{\infty}_{e}=1$ for all $e \in \vec{E}$.
We can compute $\tilde{T'}({\boldsymbol{\tilde{\eta}}^{\infty}})$
as follows:
\begin{align*}
\tilde{T'}({\boldsymbol{\tilde{\eta}}^{\infty}})
&=
 \frac{\partial  \tilde{\eta}^{t+1}_{\ed{i}{j}} }
{\partial \tilde{\eta}^{t}_{\ed{k}{l}}}
\Big{|}_{\boldsymbol{\tilde{\eta}}^{t}=1} \nonumber \\
&=
\Big(
\frac{b_{ji}(+,+)}{b_j(+)}
-
\frac{b_{ji}(-,+)}{b_j(-)}
\Big)
\mathcal{M}_{\ed{i}{j},\ed{k}{l}} \nonumber \\
&=
\frac{\chi_{ij}-m_i m_j}{1-m_j^{2}}
\mathcal{M}_{\ed{i}{j},\ed{k}{l}}.
\end{align*}
\end{proof}

\section{Idea and proof of Theorem \ref{thmindexsum}}
\subsection{Idea of Theorem \ref{thmindexsum}}
\begin{figure}
\begin{minipage}{.5\linewidth}
\begin{center}
\includegraphics[scale=0.3]{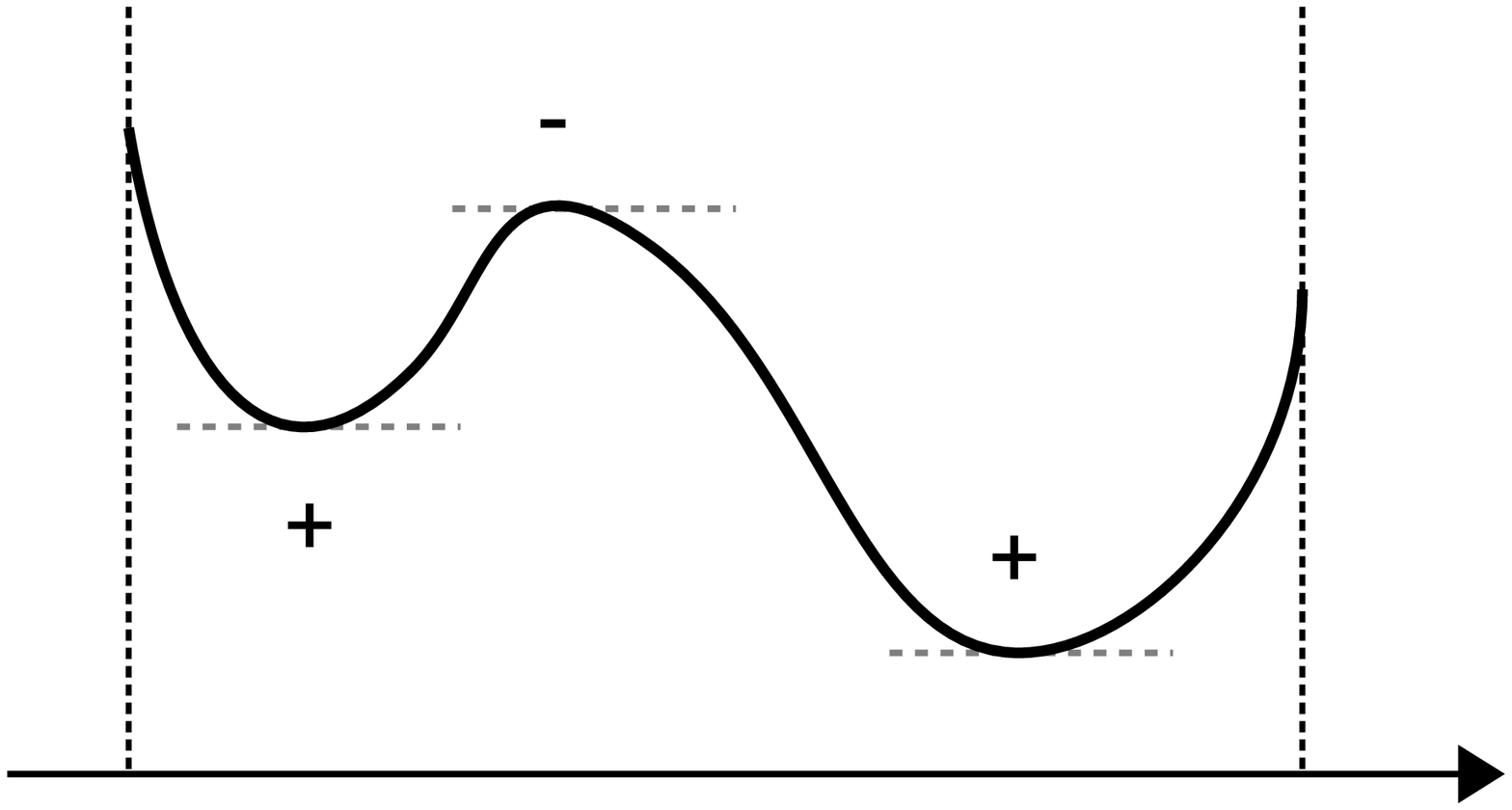} 
\caption{The sum of indexes is one. \label{figureExampleIndexFormula1}}
\end{center}
\end{minipage}
\begin{minipage}{.5\linewidth}
\begin{center}
\includegraphics[scale=0.3]{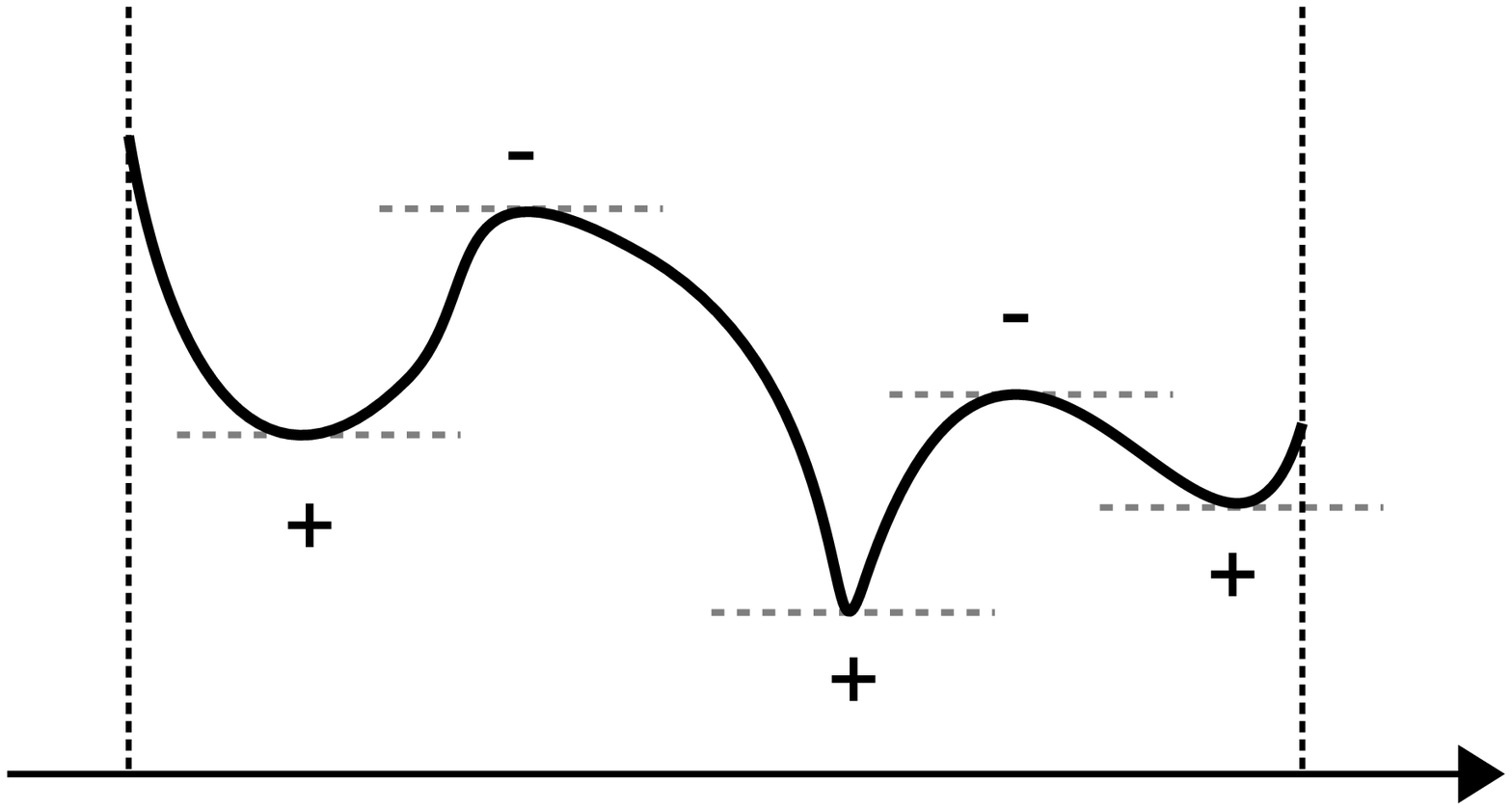} 
\caption{The sum of indexes is still one. \label{figureExampleIndexFormula2}}
\end{center}
\end{minipage}
\end{figure}

In Theorem \ref{thmindexsum}, 
we show that the sum of indexes is equal to one.
This is not so special.
The simplest example that illustrate the idea of the theorem is sketched
in figure \ref{figureExampleIndexFormula1}.
For each stationary point,
plus or minus sign is assigned depending on the sign of 
the second derivative.
When we deform the function,
the sum is still equal to one
as long as the outward gradients are positive at the boundaries.
(See figure \ref{figureExampleIndexFormula2}.)

Lemma 1, combined with Lemma 2, 
describes the behavior of the Bethe free energy near the boundary
of $L(G)$.


\subsection{Proof of Lemma 1}
\begin{proof}
First, note that it is enough to prove the assertion
when $h_i=0$ and $J_{ij}=0$.

We prove by contradiction.
Assume that $\lVert \nabla F(q_n) \rVert \not \rightarrow \infty$.
Then, there exists $R > 0$ such that 
$\lVert \nabla F(q_n) \rVert \leq R$
for infinitely many $n$.
Let 
$B_0(R)$ be the closed ball of radius $R$ centered at the origin.
Taking subsequences, if necessary,
we can assume that
\begin{equation}
 \nabla F(q_n) \rightarrow 
{}^{\exists}
\binom{\boldsymbol{\xi}}{\boldsymbol{\eta}}
\in 
B_0(R), \label{eqgradFconverges}
\end{equation}  
because of the compactness of $B_0(R)$.
Let $b_{ij}^{(n)}(x_i,x_j)$ and  $b_{i}^{(n)}(x_i)$ be the pseudomarginals
 corresponding to $q_n$.
Since $q_n \rightarrow  q_{*} \in \partial L(G)$,
there exist $ij \in E$, $x_i$ and $x_j$ such that
\begin{equation*}
b_{ij}^{(n)}(x_i,x_j) \rightarrow 0.
\end{equation*}
Without loss of generality, we assume that
$x_i=+1$ and $x_j=+1$.
From (\ref{eqgradFconverges}), we have
\begin{equation}
\nabla F(q_n)_{ij}
=
\frac{1}{4}
\log
\frac{b_{ij}^{(n)}(+,+)b_{ij}^{(n)}(-,-)}
{b_{ij}^{(n)}(+,-)b_{ij}^{(n)}(-,+)}
\longrightarrow
 \eta_{ij}. \label{eqgradFconvergesij}
\end{equation}
Therefore
$b_{ij}^{(n)}(+,-) \rightarrow 0$ or 
$b_{ij}^{(n)}(-,+) \rightarrow 0$ holds;
we assume $b_{ij}^{(n)}(+,-) \rightarrow 0$
without loss of generality.
Now we have 
\begin{equation*}
b_{i}^{(n)}(+)=b_{ij}^{(n)}(+,-)+b_{ij}^{(n)}(+,+) \rightarrow 0.
\end{equation*}

In this situation, the following claim holds.
\begin{claim}
Let $k \in N_i$.
In the limit of $n \rightarrow \infty$,
\begin{equation}
\sum_{x_i,x_k=\pm 1}x_i \log \frac{b^{(n)}_{ik}(x_i,x_k)}{b^{(n)}_i(x_i)}
=
\log
\left[
\frac{b_{ik}^{(n)}(+,+)b_{ik}^{(n)}(+,-)b_{i}^{(n)}(-)^2}
{b_{ik}^{(n)}(-,+)b_{ik}^{(n)}(-,-)b_{i}^{(n)}(+)^2}
\right] \label{eqclaim1}
\end{equation}
converges to a finite value.
\end{claim}
\begin{proof}[proof of claim]
From $b_{i}^{(n)}(+)  \rightarrow 0$, we have 
\begin{equation*}
b_{ik}^{(n)}(+,-), b_{ik}^{(n)}(+,+) \longrightarrow 0
\quad \text{ and } \quad
b_{i}^{(n)}(-)  \rightarrow 1.
\end{equation*}
\textbf{Case 1:}
$b_{ik}^{(n)}(-,+) \longrightarrow b_{ik}^{*}(-,+) \neq 0$ and
$b_{ik}^{(n)}(-,-) \longrightarrow b_{ik}^{*}(-,-) \neq 0$. \\
In the same way as (\ref{eqgradFconvergesij}),
\begin{equation*}
\nabla F(q_n)_{ik}
=
\frac{1}{4}
\log
\frac{b_{ik}^{(n)}(+,+)b_{ik}^{(n)}(-,-)}
{b_{ik}^{(n)}(+,-)b_{ik}^{(n)}(-,+)}
\longrightarrow
 \eta_{ik}. 
\end{equation*}
Therefore
\begin{equation*}
\frac{b_{ik}^{(n)}(+,+)}
{b_{ik}^{(n)}(+,-)}
\longrightarrow
{}^{\exists}r \neq 0.
\end{equation*}
Then we see that (\ref{eqclaim1}) converges to a finite value.

\textbf{Case 2:}
$b_{ik}^{(n)}(-,+) \longrightarrow 1$ and
$b_{ik}^{(n)}(-,-) \longrightarrow 0$. \\
Similar to the case 1, we have
\begin{equation*}
\frac{b_{ik}^{(n)}(+,+)b_{ik}^{(n)}(-,-)}
{b_{ik}^{(n)}(+,-)}
\longrightarrow
{}^{\exists}r \neq 0.
\end{equation*}
Therefore
$\frac{b_{ik}^{(n)}(+,-)}{b_{ik}^{(n)}(+,+)} \rightarrow 0$.
This implies that
$\frac{b_{i}^{(n)}(+)}{b_{ik}^{(n)}(+,+)} \rightarrow 1$.
Then we see that (\ref{eqclaim1}) converges to a finite value.

\textbf{Case 3:}
$b_{ik}^{(n)}(-,+) \longrightarrow 0$ and
$b_{ik}^{(n)}(-,-) \longrightarrow 1$. \\
Same as the case 2.
\end{proof}

Now let us get back to the proof of Lemma 1.
We rewrite (\ref{Bethem}) as
\begin{equation}
\nabla F(q_n)_{i}
=
\frac{1}{2}\log b^{(n)}_i(+)
-
\frac{1}{2}\log b^{(n)}_i(-)
+ \frac{1}{4}\sum_{k \in N_i}\sum_{x_i,x_k=\pm 1}x_i \log
\frac{b^{(n)}_{ik}(x_i,x_k)}{b^{(n)}_{i}(x_i)} \label{lemma1finaleq}
\end{equation}
From (\ref{eqgradFconverges}), this value converges to $\xi_i$.
The second and the third terms in (\ref{lemma1finaleq}) 
converges to a finite value, while the
 first value converges to infinite.
This is a contradiction.
\end{proof}

\subsection{Detailed proof of Theorem \ref{thmindexsum}}
\begin{proof}
Define a map $\Phi:L(G) \rightarrow \mathbb{R}^{N+M}$ by
\begin{align}
\Phi(q)_{i}& = 
(1-d_i) \frac{1}{2} \sum_{x_i = \pm 1} x_i \log b_i(x_i)
+ \frac{1}{4}\sum_{k \in N_i}\sum_{x_i,x_k=\pm 1}x_i \log
 b_{ik}(x_i,x_k), \label{Phii} \\
\Phi(q)_{ij}& = 
\frac{1}{4} \sum_{x_i,x_j=\pm 1}x_i x_j \log  b_{ij}(x_i,x_j), \label{Phiij}
\end{align}
where $b_{ij}(x_i,x_j)$ and $b_i(x_i)$ 
are given by $q=\{m_i,\chi_{ij}\} \in L(G)$.
Therefore, 
we have $\nabla F = \Phi - \binom{\boldsymbol{h}}{\boldsymbol{J}}$
and
$ \nabla \Phi =  \nabla^{2} F$.
Then following claim holds. 
\begin{claim}
The sets 
$\Phi^{-1}(\binom{\boldsymbol{h}}{\boldsymbol{J}}),\Phi^{-1}(0) \subset L(G)$ 
are finite and 
\begin{equation}
\sum_{q \in \Phi^{-1}
({h \atop J} ) 
}
\sgn( \det \nabla \Phi(q))
=
\sum_{q \in \Phi^{-1}
(0) 
}
\sgn( \det \nabla \Phi(q)), \label{eq1thmindexsum}
\end{equation}
holds.
\end{claim}
Before the proof of this claim, 
we prove Theorem \ref{thmindexsum} under the claim.

From (\ref{Phii}) and (\ref{Phiij}),
it is easy to see that 
$\Phi(q)=0 \Leftrightarrow q=\{m_i=0,\chi_{ij}=0\}$.
At this point,
we can easily check that
$\nabla \Phi =  \nabla^{2} F$ is a positive definite matrix.
Therefore the right hand side of (\ref{eq1thmindexsum}) is
equal to one.
The left hand side of (\ref{eq1thmindexsum})
is equal to the left hand side of (\ref{thmindexsumeq1}),
because 
$q \in \Phi^{-1}(\binom{\boldsymbol{h}}{\boldsymbol{J}} )  \Leftrightarrow \nabla F(q)=0$.
Then the assertion of Theorem \ref{thmindexsum} is proved.
\end{proof}

\begin{proof}[Proof of the claim]
First, we prove that 
$\Phi^{-1}( \binom{\boldsymbol{h}}{\boldsymbol{J}}  )=(\nabla F)^{-1}(0)$ is a finite set.
If not, we can choose a sequence $\{q_n\}$ of distinct points 
from this set.
Let $\overline{L(G)}$
be the closure of $L(G)$.
Since $\overline{L(G)}$ is compact, we can choose a subsequence that
 converges to some point $q_{*} \in \overline{L(G)}$. 
From Lemma 1, $q_{*} \in L(G)$ and $\nabla F(q_{*})=0$ hold.
By the assumption in Theorem \ref{thmindexsum},
we have $\det\nabla^{2} F(q_{*})\neq 0$.
This implies that $\nabla F(q)\neq 0$ in some neighborhood of $q_{*}$.
This is a contradiction because $q_n \rightarrow q_{*}$.

Secondly, we prove the equality (\ref{eq1thmindexsum}) using Lemma 2.
Define a sequence of compact convex sets
$C_n:=\{q \in L(G)|\sum_{ij \in E}\sum_{x_i,x_j}-\log b_{ij} \leq n \}$,
which increasingly converges to $L(G)$.
Since $\Phi^{-1}(0)$ and $\Phi^{-1} \binom{\boldsymbol{h}}{\boldsymbol{J}}$
are finite,
they are included in $C_n$ for sufficiently large $n$.
Take $K>0$ and $\epsilon >0$ to satisfy 
$K -\epsilon > \lVert \binom{\boldsymbol{h}}{\boldsymbol{J}}  \rVert$.
From Lemma 1,
we see that $\Phi(\partial C_{n}) \cap B_0(K)=\phi$
for sufficiently large $n$.
Let $n_o$ be such a large number.
Let $\Pi_{\epsilon}:\mathbb{R}^{N+M} \rightarrow  B_0(K)$ be a smooth
 map that is identity on $B_0(K-\epsilon)$, 
monotonically increasing on $\lVert x \rVert$,
and
 $\Pi_{\epsilon}(x)=\frac{K}{\lVert x \rVert}x$
for $\lVert x \rVert \geq K$.
Then we obtain a composition map 
$\tilde{\Phi }:=\Pi_{\epsilon} \circ \Phi :C_{n_0} \rightarrow  B_0(K)$
that satisfy $\tilde{\Phi }(\partial C_{n_0}) \subset \partial B_0(K)$.
By definition, we have
$\Phi^{-1}(0)=\tilde{\Phi}^{-1}(0)$ and 
$\Phi^{-1} \binom{\boldsymbol{h}}{\boldsymbol{J}}
= \tilde{\Phi}^{-1} \binom{\boldsymbol{h}}{\boldsymbol{J}}$.
Therefore, both $0$ and $\binom{\boldsymbol{h}}{\boldsymbol{J}}$ are regular values
 of $\tilde{\Phi}$.
From Lemma 2, we have 
\begin{equation*}
\sum_{q \in \tilde{\Phi}^{-1}
({h \atop J} ) 
}
\sgn( \det \nabla \tilde{\Phi}(q))
=
\sum_{q \in \tilde{\Phi}^{-1}
(0) 
}
\sgn( \det \nabla \tilde{\Phi}(q)).
\end{equation*}
Then, the assertion of the claim is proved.
\end{proof}

\section{Proof of Corollary \ref{cor2loop}}
\subsection{Detailed proof of Example 2}
\begin{figure}
\begin{minipage}{.5\linewidth}
\begin{center}
\includegraphics[scale=0.4]{inputfiles/ExampleGraph1.eps} 
\caption{The graph $G$. \label{figureExampleGraph1sup}}
\end{center}
\end{minipage}
\begin{minipage}{.5\linewidth}
\begin{center}
\includegraphics[scale=0.3]{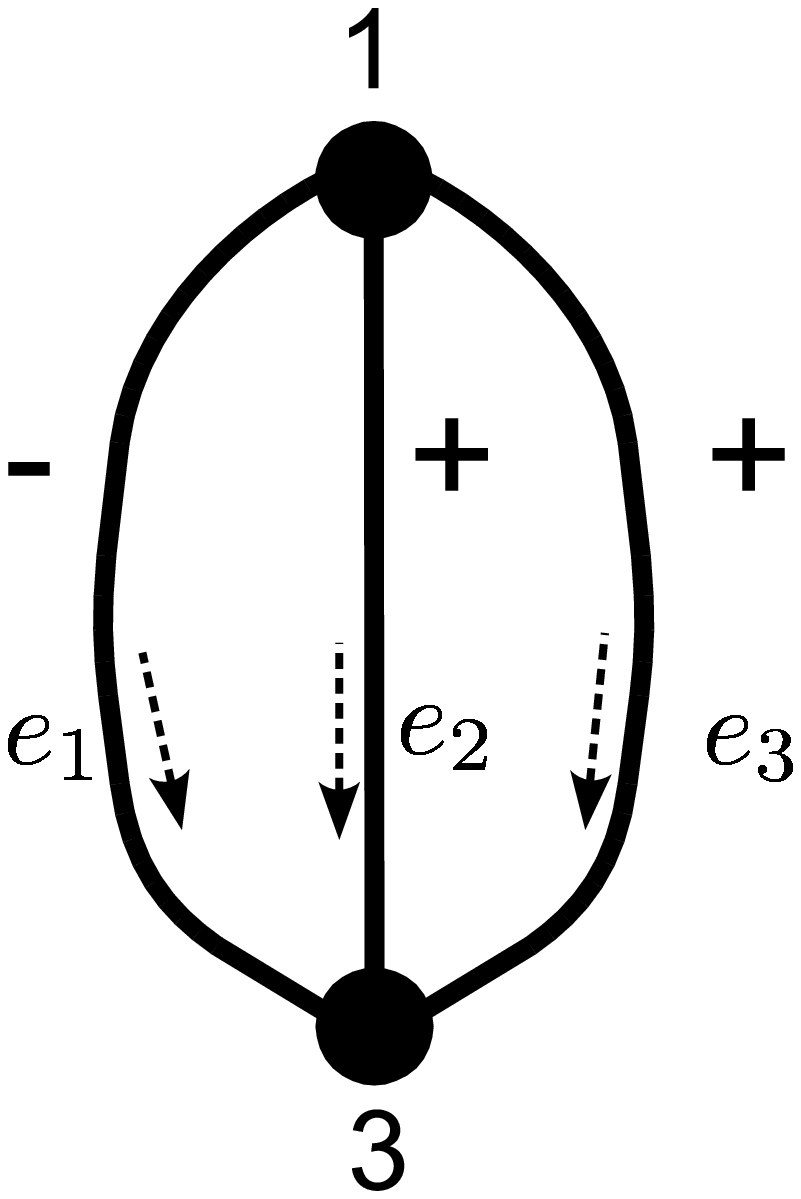} 
\caption{The graph $\cg{G}$. \label{figureExampleGraph2sup}}
\end{center}
\end{minipage}
\end{figure}

In this subsection we prove the assertion of Corollary \ref{cor2loop} for the
graph of Example 2, which is 
displayed in figure \ref{figureExampleGraph1sup}.
The $+$ and $-$ signs represent that of two body interactions.

It is enough to check that
$\det(I -  \mathcal{B}\mathcal{M}) > 0$
for arbitrary
$0 \leq \beta_{13},\beta_{23},\beta_{14},\beta_{34} < 1$
and $-1 < \beta_{12} \leq 0$.
The graph $\cg{G}$ in figure \ref{figureExampleGraph2sup}
is obtained by erasing vertices $2$ and $4$ in $G$.
To compute $\det(I -  \mathcal{B}\mathcal{M})$,
it is enough to consider $\cg{G}$.
In fact 
\begin{align}
\det(I -  \mathcal{B}\mathcal{M})
&=
\zeta_{G}(\boldsymbol{\beta})^{-1} \nonumber \\
&=
\prod_{\mathfrak{p} \in P }
(1-g(\mathfrak{p}))              \label{GtoG'eq2}  \\
&=
\prod_{\mathfrak{\cg{p}} \in \cg{P} }
(1-g(\mathfrak{\cg{p}}))             \label{GtoG'eq3}   \\
&=
\zeta_{\cg{G}}(\boldsymbol{\cg{\beta}})^{-1} 
=
\det(I -  \mathcal{\cg{B}}\mathcal{\cg{M}}), \nonumber
\end{align}
where $\cg{\beta}_{e_1}:=\beta_{12}\beta_{23}$,
$\cg{\beta}_{e_2}:=\beta_{13}$, 
$\cg{\beta}_{e_3}:=\beta_{14}\beta_{34}$ and
$\cg{\beta}_{e_i}=\cg{\beta}_{\bar{e_i}}$.
The equality between (\ref{GtoG'eq2}) and (\ref{GtoG'eq3})
is obtained by the one to one correspondence between
prime cycles of $G$ and $\cg{G}$.

By definition,
we have 
\begin{equation*}
\mathcal{\cg{B}}\mathcal{\cg{M}}
=
\left[
\begin{array}{cccccc}
0 & 0 & 0 & 0   & \cg{\beta}_{e_1} & \cg{\beta}_{e_1}   \\
0       & 0 &0 & \cg{\beta}_{e_2}  & 0 & \cg{\beta}_{e_2} \\
0 & 0  & 0 &  \cg{\beta}_{e_3}  & \cg{\beta}_{e_3} & 0         \\ 
0 & \cg{\beta}_{e_1}& \cg{\beta}_{e_1} &0 & 0 & 0  \\
\cg{\beta}_{e_2} & 0& \cg{\beta}_{e_2} &0 & 0 & 0  \\
\cg{\beta}_{e_3} & \cg{\beta}_{e_3}& 0 & 0 & 0 & 0 
\end{array}
\right],
\end{equation*}
where
the rows and columns are indexed by $e_1,e_2,e_3,\bar{e_1},\bar{e_2}$
and $\bar{e_3}$.
Then the determinant is
\begin{align*}
\det(I -  \mathcal{\cg{B}}\mathcal{\cg{M}})
&=
\det \left[
I -
\left(
\begin{array}{ccc}
 0   & \cg{\beta}_{e_1} & \cg{\beta}_{e_1}    \\
 \cg{\beta}_{e_2}  & 0 & \cg{\beta}_{e_2} \\
  \cg{\beta}_{e_3}  & \cg{\beta}_{e_3} & 0         
\end{array}
\right)
\right]
\det \left[
I +
\left(
\begin{array}{ccc}
 0   & \cg{\beta}_{e_1} & \cg{\beta}_{e_1}    \\
 \cg{\beta}_{e_2}  & 0 & \cg{\beta}_{e_2} \\
  \cg{\beta}_{e_3}  & \cg{\beta}_{e_3} & 0         
\end{array}
\right)
\right] \nonumber \\
&=
(1 - \cg{\beta}_{e_1} \cg{\beta}_{e_2} - \cg{\beta}_{e_1} \cg{\beta}_{e_3} -
 \cg{\beta}_{e_2} \cg{\beta}_{e_3} 
- 2 \cg{\beta}_{e_1} \cg{\beta}_{e_2} \cg{\beta}_{e_3}) \nonumber \\
& \quad \qquad
(1 - \cg{\beta}_{e_1} \cg{\beta}_{e_2} - \cg{\beta}_{e_1} \cg{\beta}_{e_3} - 
\cg{\beta}_{e_2} \cg{\beta}_{e_3}
+ 2 \cg{\beta}_{e_1} \cg{\beta}_{e_2} \cg{\beta}_{e_3}).
\end{align*}
Since  $-1 < \cg{\beta}_{e_1} \leq 0$ and
$0 \leq \cg{\beta}_{e_2},\cg{\beta}_{e_3} < 1$,
we conclude that this is positive.

\subsection{Other cases}
\begin{figure}
\begin{center}
\includegraphics[scale=0.4]{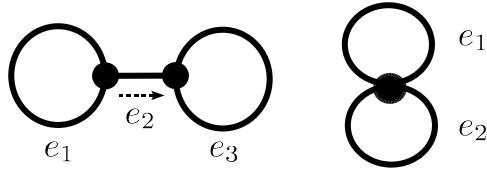} 
\caption{Two other types of graphs. \label{figureExampleGraph3s}}
\end{center}
\end{figure}

There are two operations on graphs
that do not change the set of prime cycles. 
The first one is adding or removing a vertex of degree two on any edge.
The second one is adding or removing an edge with a vertex of degree
one.
With these two operations,
all graphs that have two linearly independent cycles 
are reduced to three types of graphs. 
The first type is in figure \ref{figureExampleGraph2sup}.
The other types are in figure \ref{figureExampleGraph3s}.

\begin{figure}
\includegraphics[scale=0.4]{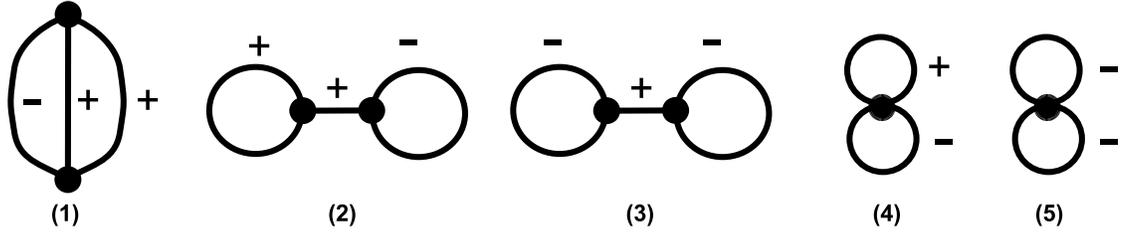} 
\caption{List of interaction types. \label{figureExampleGraphAllTypes}}
\end{figure}

Up to equivalence of interactions,
all types of signs of two body 
interactions are listed in figure 
\ref{figureExampleGraphAllTypes} except for the attractive case.
We check the uniqueness for each case in order.

\textbf{Case (1):} Proved in Example 2.

\textbf{Case (2):}
In this case,
\begin{equation*}
\mathcal{B}\mathcal{M}
=
\left[
\begin{array}{cccccc}
 {\beta}_{e_1} &0   & 0 &0   &  {\beta}_{e_1} & 0   \\
{\beta}_{e_2}      & 0 &0 & {\beta}_{e_2}  & 0 & 0 \\
0 & {\beta}_{e_3}  &  {\beta}_{e_3} &  0  & 0 & 0         \\ 
0 & 0 & 0 &{\beta}_{e_1} & {\beta}_{e_1} & 0  \\
0 & 0& {\beta}_{e_2} &0 & 0 & {\beta}_{e_2}  \\
0 &  {\beta}_{e_3} & 0 & 0 & 0 & {\beta}_{e_3} 
\end{array}
\right],
\end{equation*}
where rows and columns are labeled by
$e_{1},e_{2},e_{3},\bar{e_1},\bar{e_2}$ and $\bar{e_3}$.
Then the determinant is
\begin{equation}
\det(I -  \mathcal{B}\mathcal{M})
=
(1- {\beta}_{e_1})
(1- {\beta}_{e_3})
(1- {\beta}_{e_1}-{\beta}_{e_3}+ {\beta}_{e_1}{\beta}_{e_3} 
-4 {\beta}_{e_1}{\beta}_{e_2}^2 {\beta}_{e_3} ).    \label{zetatype2}
\end{equation}

This is positive when 
$0 \leq \beta_{e_1},{\beta}_{e_2} < 1$ and
$-1 < \beta_{e_3} \leq 0$.

\textbf{Case (3):}
The determinant (\ref{zetatype2})
is also positive when 
$0 \leq {\beta}_{e_2} < 1$ and
$-1 < \beta_{e_1} ,\beta_{e_3} \leq 0$.

\textbf{Case (4):}
In this case,
\begin{equation*}
\mathcal{B}\mathcal{M}
=
\left[
\begin{array}{cccc}
 {\beta}_{e_1} &  {\beta}_{e_1}   &  0 & {\beta}_{e_1}      \\
{\beta}_{e_2}      & {\beta}_{e_2}    & {\beta}_{e_2}    & 0   \\
0 & {\beta}_{e_1}  &  {\beta}_{e_1} &  {\beta}_{e_1}          \\ 
{\beta}_{e_2}  & 0  & {\beta}_{e_2}  &{\beta}_{e_2}  
\end{array}
\right],
\end{equation*}
where rows and columns are labeled by
$e_{1},e_{2},\bar{e_1}$ and $\bar{e_2}$.
Then we have
\begin{equation}
\det(I -  \mathcal{B}\mathcal{M})
=
(1- {\beta}_{e_1})
(1- {\beta}_{e_2})
(1- {\beta}_{e_1}-{\beta}_{e_2}
- 3{\beta}_{e_1}{\beta}_{e_2})
.    \label{zetatype3}
\end{equation}
This is positive when 
$0 \leq {\beta}_{e_1} < 1$ and
$-1 < \beta_{e_2} \leq 0$.

\textbf{Case (5):}
The determinant (\ref{zetatype3}) is positive
when
$-1 < \beta_{e_1},\beta_{e_2} \leq 0$.